\documentclass{l4dc2027}

\title[Certify, Then Select]{Certify, Then Select: Viability-Aware Runtime Steering\\ of Frozen Diffusion Policies}
\usepackage{times}
\usepackage{booktabs}
\usepackage{longtable}
\usepackage{multirow}
\usepackage{xspace}
\usepackage{bm}
\newtheorem{assumption}{Assumption}
\newcommand{\E}{\mathbb E}
\newcommand{\Prob}{\mathbb P}
\newcommand{\KL}{\mathrm{KL}}
\newcommand{\cF}{\mathcal F}
\newcommand{\cN}{\mathcal N}
\newcommand{\ind}{\mathbf 1}
\newcommand{\miss}{\textsc{cast}\xspace}
\newcommand{\secref}[1]{Section~\ref{#1}}
\newcommand{\thmref}[1]{Theorem~\ref{#1}}
\newcommand{\propref}[1]{Proposition~\ref{#1}}
\newcommand{\tabref}[1]{Table~\ref{#1}}
\newcommand{\figref}[1]{Figure~\ref{#1}}

\newcommand{\LatMissCpu}{15\xspace}

\newcommand{\CdSquarePillarFallback}{55.7\xspace}
\newcommand{\CdCanFreeA}{0.560\xspace}
\newcommand{\CdCanFreeGreedy}{0.905\xspace}

\newcommand{\SpState}{0.051\xspace}
\newcommand{\SpCam}{0.013\xspace}
\newcommand{\OrCam}{0.75\xspace}

\newcommand{\HlCanCast}{0.88\xspace}

\newcommand{\HlCanStf}{0.88\xspace}

\newcommand{\HlBudCast}{0.68\xspace}

\newcommand{\HlBudStf}{0.46\xspace}

\newcommand{\HlBudScale}{0.42\xspace}

\newcommand{\RgFilterLo}{+0.22\xspace}
\newcommand{\RgFilterHi}{+0.31\xspace}

\newcommand{\RgStfLargePool}{+0.04 [+0.01, +0.08]\xspace}

\newcommand{\RgBudScaleQ}{0.46\xspace}
\newcommand{\RgBudCastQ}{0.71\xspace}

\newcommand{\RgBudCastB}{0.53\xspace}
\newcommand{\RgBudStfB}{0.16\xspace}

\newcommand{\RgBudScaleB}{0.26\xspace}

\newcommand{\RgNTwo}{0.87\xspace}

\newcommand{\RgNFour}{0.89\xspace}

\newcommand{\RgNThirtyTwo}{0.77\xspace}

\newcommand{\ZfCast}{0.98\xspace}

\newcommand{\ZfStallFilter}{20\xspace}
\newcommand{\ZfPsf}{0.75\xspace}

\newcommand{\ZfTask}{0.81\xspace}
\newcommand{\ZfStallTask}{15\xspace}

\newcommand{\ZfStfTask}{0.86\xspace}

\newcommand{\ZfStfReq}{0.98\xspace}

\newcommand{\ZfN}{500\xspace}

\newcommand{\QzFilter}{0.44\xspace}
\newcommand{\QzFailFilter}{56\xspace}
\newcommand{\QzMppi}{0.61\xspace}

\newcommand{\QzTask}{0.68\xspace}

\newcommand{\QzReq}{1.00\xspace}

\newcommand{\QCriticRtwo}{0.51--0.63\xspace}
\newcommand{\QReqAuc}{0.99\xspace}
\newcommand{\CvEight}{0.99\xspace}
\newcommand{\CvStallEight}{0.4\xspace}
\newcommand{\CvFour}{0.80\xspace}

\newcommand{\CvTwo}{0.83\xspace}

\newcommand{\CvOne}{0.82\xspace}

\newcommand{\CvVonly}{0.99\xspace}

\newcommand{\QvSixteen}{0.93\xspace}

\newcommand{\HbFellFilter}{74\xspace}

\newcommand{\CmUni}{0.14\xspace}

\newcommand{\RvSwLo}{2\xspace}
\newcommand{\RvSwHi}{5\xspace}

\newcommand{\RvSwPbLo}{14\xspace}
\newcommand{\RvSwPbHi}{20\xspace}

\newcommand{\SbCastStf}{$-$0.04 [$-$0.07, $-$0.00]\xspace}

\newcommand{\VbCastFilter}{+0.08 [+0.03, +0.14]\xspace}

\newcommand{\PedVl}{0.95\xspace}
\newcommand{\PedLogCrit}{0.98\xspace}
\newcommand{\PedVlFilter}{+0.23 [+0.19, +0.26]\xspace}
\newcommand{\PedVlMppi}{+0.12 [+0.08, +0.15]\xspace}
\newcommand{\PedVlCastv}{+0.13 [+0.10, +0.16]\xspace}
\newcommand{\MfUni}{0.95\xspace}

\newcommand{\MfQuad}{0.98\xspace}

\newcommand{\MfMmGap}{$-$0.01 [$-$0.02, +0.00]\xspace}

\newcommand{\MfViolMb}{6.4\xspace}
\newcommand{\MfViolMf}{1.6\xspace}
\newcommand{\MfViolCast}{38\xspace}

\newcommand{\MfAucLo}{0.975\xspace}

\newcommand{\CtlAuc}{0.98\xspace}

\newcommand{\CtlVal}{0.97\xspace}

\newcommand{\EstEpsB}{0.42\xspace}

\newcommand{\EstErrA}{0.07\,\%\xspace}

\newcommand{\EstTable}{Decisions with a certified candidate of reference viability $\ge0.9$: 16\,901; the executed chunk had reference viability below 0.1 in 0.07\,\% of them (bound $\varepsilon_4=0.42$). For $(v_{\rm lo},v_{\rm hi})=(0.25,0.75)$: 0.26\,\% of 17\,057 (bound vacuous, 2.14). Only 4.96\,\% of 140\,200 certified candidates have reference viability between 0.1 and 0.9: viability is nearly always close to 0 or 1, where $M=4$ suffices.\xspace}

\newcommand{\SqFbAllN}{554\xspace}

\newcommand{\SqFbN}{1\,200\xspace}

\newcommand{\DeVl}{0.89\xspace}
\newcommand{\DeSc}{0.76\xspace}

\newcommand{\DeDiffK}{$-$0.01 [$-$0.03, $-$0.00]\xspace}
\newcommand{\DeDiffT}{+0.01 [$-$0.00, +0.03]\xspace}

\newcommand{\BkEpisodesAll}{30\,000\xspace}
\newcommand{\BkEdgeUncert}{76\xspace}
\newcommand{\BkEdgeN}{76\xspace}

\newcommand{\BkAfterE}{50\xspace}
\newcommand{\BkInS}{48\xspace}

\newcommand{\BkFilterGE}{+0.15 [+0.11, +0.18]\xspace}

\newcommand{\BkCastJEo}{0.60\xspace}
\newcommand{\BkCastJEb}{0.81\xspace}

\newcommand{\BkCastGE}{+0.21 [+0.18, +0.24]\xspace}

\newcommand{\BkCastvGE}{+0.02 [+0.01, +0.03]\xspace}

\providecommand{\CdCanObstRankPt}{$+0.32$\xspace}

\providecommand{\CdCanPillarRankPt}{$+0.34$\xspace}

\providecommand{\CdSquareObstRankPt}{$+0.16$\xspace}

\providecommand{\CdSquarePillarRankPt}{$+0.03$\xspace}

\providecommand{\CdCanFreeAR}{0.56\xspace}
\providecommand{\CdCanFreeGreedyR}{0.91\xspace}

\providecommand{\BuGreedySTFR}{+0.22 [+0.16, +0.29]\xspace}

\providecommand{\TauGapOnePt}{$+0.10$\xspace}

\providecommand{\HlCanGainPt}{$+0.32$\xspace}

\providecommand{\HlSqGainPt}{$+0.09$\xspace}

\providecommand{\RgBudPoolPt}{$+0.26$\xspace}

\providecommand{\RgNDropPt}{$-0.10$\xspace}

\providecommand{\ZfTargetGainPt}{$+0.16$\xspace}

\providecommand{\QbStfPt}{$+0.09$\xspace}
\providecommand{\QbFilterPt}{$+0.36$\xspace}

\providecommand{\CvTaskPt}{$+0.18$\xspace}
\providecommand{\CvStfPt}{$+0.13$\xspace}

\providecommand{\QvTaskPt}{$+0.25$\xspace}

\providecommand{\HbFilterPt}{$+0.53$\xspace}

\providecommand{\QbvCastPt}{$+0.02$\xspace}

\providecommand{\HbvCastPt}{$+0.09$\xspace}

\providecommand{\SbCastFilterPt}{$+0.08$\xspace}

\providecommand{\VrNinePt}{$+0.22$\xspace}

\providecommand{\MfUniGainPt}{$+0.14$\xspace}

\author{%
 \Name{Angelo Moroncelli} \Email{angelo.moroncelli@supsi.ch}\\
 \addr SUPSI-IDSIA, USI, Lugano, Switzerland
 \AND
 \Name{Loris Roveda} \Email{loris.roveda@supsi.ch}\\
 \addr SUPSI-IDSIA, Lugano, Switzerland; Politecnico di Milano, Milan, Italy%
}

\begin{document}
\maketitle
\begin{abstract}%
Diffusion policies, which generate short action sequences by denoising random noise, are a leading way to learn robot skills from demonstrations. Once deployed, they meet constraints the demonstrations never contained, and retraining for each is impractical. A safety filter edits the policy's action to satisfy them, but the edit ignores the task, and a large edit can leave the policy where it cannot continue. We propose \textbf{\textit{CAST}} (\textbf{\textit{C}}ertified \textbf{\textit{A}}ction \textbf{\textit{S}}election at \textbf{\textit{T}}est time): \textbf{draw} a few action chunks from the frozen policy, \textbf{certify} each with a barrier condition, and \textbf{execute} the certified chunk that a learned value ensemble ranks highest. We prove that a certified step is safe whatever the critic does; that, whenever a certified sample exists, the executed chunk stays within $\log N$ nats of the policy conditioned on feasibility, while a safety-filter projection can be singular to it; and that an exact closed-loop identity splits the gain of ranking into availability, spread and the critic's target. Certifying the next chunk is not enough: a generative controller can have safe actions available now and none later. We characterise the safe control authority of a frozen policy under finite sampling, and \miss-V preserves it through a policy-conditioned viability, estimated with a model under a finite-sample guarantee or learned from the controller's own logs (\miss-VL). On robomimic manipulation, a planar quadrotor, a MuJoCo humanoid and a unicycle, with frozen policies, \miss raises joint task-and-constraint success above the output filter in nearly every setting; certifying before ranking is decisive where a filter's edit is large, and viability removes most dead ends that one-step certification cannot see. Project page: \url{https://clmoro.github.io/cast}.
\end{abstract}


\section{Introduction}\label{sec:intro}
Diffusion policies learn dexterous manipulation from demonstrations~\citep{chi2023diffusion}. Once deployed, a policy meets constraints its demonstrations never contained: a person reaches across the table, a shelf is mounted above the workspace, the commanded effort must stay within a budget. Retraining for each is impractical. We want to enforce a constraint that arrives at run time, a \emph{request}, on a policy that stays frozen.

\begin{figure}[t]
\centering
\includegraphics[width=\textwidth]{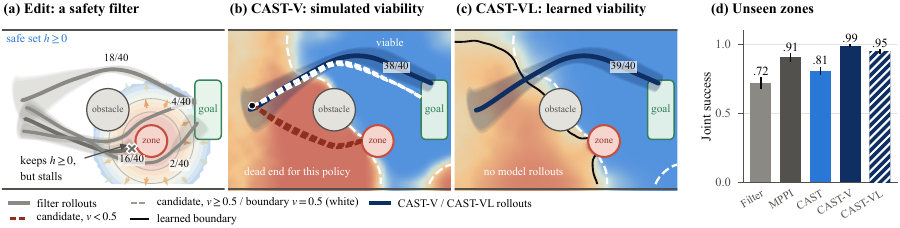}
\caption{\textbf{Edit or select.} A keep-out zone, unseen in training, blocks one of the two routes of a frozen unicycle policy. (a) A safety filter keeps the robot in the safe set $h\geq0$ (shading: the barrier), but its rollouts on the blocked route stall at the zone: the policy never learned to continue from there (40 rollouts, with counts). (b) Viability $v$ of the policy's certified chunks, by simulation: the probability that it keeps sampling certified chunks over the next 8 decisions. Certified chunks exist in the red region too, but lead to a dead end; of the $N=8$ drawn at one decision four head into it, and \miss-V executes a viable one. (c) The same map from a network trained on the controller's logs, without model rollouts (\miss-VL): its boundary $v=0.5$ (black) follows the simulated one (white) where the rollouts pass. (d) Joint success on \ZfN{} unseen zones per policy; \miss, \miss-V and \miss-VL use the task critic, MPPI the model, map and goal.}
\label{fig:concept}
\end{figure}

The classical tool is a safety filter: take the action the policy proposes and edit it as little as needed for a certificate, such as a control barrier function, to hold~\citep{ames2017cbfqp,wabersich2021predictive,hsu2023safetyfilter}. Variants for diffusion policies apply the same edit inside the denoising loop~\citep{xiao2023safediffuser,romer2025dpcc}. An edit answers the question ``is this action safe?''. It does not ask whether the task can still be completed. A small edit leaves the policy as good or as bad as it was. A large one, as under an effort budget or when a zone blocks the policy's route, leaves the robot where the policy has no data, and it stalls (\figref{fig:concept}a). Neither helps the task: in our experiments no method that keeps the policy and has no critic reliably beats the plain output filter in joint task-and-request success (\tabref{tab:main}).

A diffusion policy offers a resource that a deterministic controller lacks: for one observation it can propose many different, equally plausible actions. \miss (\emph{Certified Action Selection at Test time}, Algorithm~\ref{alg:cast}) uses this resource twice. It draws a few action chunks from the frozen policy, certifies each against the request, and executes the certified chunk that a value ensemble ranks highest. Only if no sample is certified does it fall back to the filter. The certificate provides safety, the critic provides task success, and whenever a certified sample exists the executed action is one the policy itself produced. But certification looks one step ahead, and a generative controller can have safe actions available today but lack safe behaviours tomorrow. Runtime constraint steering should preserve the learned policy's future control authority, not merely certify its next action: \miss-V prefers chunks after which certified samples remain likely (\figref{fig:concept}b).

\textbf{Contributions.}
\begin{itemize}
\item \emph{Method.} \miss enforces a runtime request on a frozen diffusion policy by certifying sampled chunks and then ranking the certified ones (\secref{sec:method}). It needs no retraining and one batched sampling pass per decision (\LatMissCpu{}\,ms on Can). Select-then-filter, which uses the same samples and critic but reverses the order, isolates the benefit of certifying before ranking.
\item \emph{Theory} (\secref{sec:theory}; proofs in Appendix~\ref{app:proofs}). Four results say what each ingredient provides; the first three specialise known tools to sampling, certifying and selecting, and the fourth is the new object. \textbf{\textit{Safety:}} a certified step is safe whatever the critic does, and one conformal calibration bounds the forecast risk for every controller (\thmref{thm:safety}). \textbf{\textit{Faithfulness:}} when a certified sample exists, selection stays within $\log N$ nats of the policy conditioned on feasibility, whereas a projection can leave its support (\thmref{thm:minimal}). \textbf{\textit{Performance:}} an exact closed-loop identity says when ranking improves the task, through availability, spread and the critic's target (\thmref{thm:selection}). \textbf{\textit{Authority:}} the best any selection rule can do with $N$ samples to keep certified actions available grows with $N$ only up to what the policy samples with positive probability (\propref{prop:authority}); \miss-V and \miss-VL estimate it.
\item \emph{Evidence} (\secref{sec:experiments}). On robomimic, a quadrotor, a humanoid and a unicycle, with frozen policies and eight samples, \miss beats the output filter in nine of ten settings (\figref{fig:forest}a; the tenth, a crossing pedestrian, is a tie). The critic supplies most of this gain; certifying before ranking adds to it where the edit is large (Can, $b=0.15$: \RgBudCastB{} against \RgBudStfB{} for select-then-filter). Where a zone blocks a learned route, \miss-V restores availability (unicycle \CvEight, against \ZfTask{} for \miss), and \miss-VL does so from logs, without simulating continuations (quadrotor \MfQuad; MPPI \QzMppi).
\end{itemize}

\textbf{Related work.} Safety filters correct a nominal input~\citep{ames2017cbfqp,wabersich2021predictive,hsu2023safetyfilter,nakamura2025generalizing,nakamura2025latentcbf,hu2026vlsa}; shielding and sampling-based filters reject unsafe actions without a task model~\citep{alshiekh2018shielding,park2026sampling}. Constrained diffusion edits the denoising iterates~\citep{xiao2023safediffuser,romer2025dpcc,gadginmath2026constricting,romer2026pacs} or the input noise~\citep{chen2026inspo}. Safety critics, recovery regions and viability can be learned from data~\citep{srinivasan2020safetycritic,thananjeyan2021recovery,zheng2024fisor,wang2026movil}, and planning over random action sets is classical~\citep{boutilier2018stochastic}; \miss-V asks whether a fixed generative controller will keep sampling certified actions under a request it never saw. Sample-and-rank methods choose among policy samples by value~\citep{fujimoto2019bcq,nakamoto2024vgps}, density~\citep{rosasco2025kdpe} or a verifier~\citep{kwok2025robomonkey}, without certificates. We analyse how sampling, rejection and ranking interact in frozen generative control, including when they cannot help, with information projection~\citep{csiszar1975idivergence}, performance differences~\citep{kakade2002approximately} and conformal prediction~\citep{lindemann2023conformal,cleaveland2024conformal}.

\section{Problem}\label{sec:problem}
\textbf{Policy.} At decision $k$ a frozen generative policy maps the observation $o_k$ and a latent $x\sim\cN(0,I)$ to a chunk $a=G(o_k,x)$ of commands, whose first $T$ are executed in open loop; the law of $G(o_k,x)$ is the policy $\pi(\cdot\mid o_k)$, which we may sample but not change.

\textbf{Requests.} A request constrains a robot quantity $p_t$ (the hand position, or the speed) against a hazard with state $\xi_t$: it holds at time $t$ if $h(p_t,\xi_t)\ge0$, where the barrier $h$ is 1-Lipschitz in each argument, as a signed distance is. A budget asks that the average command effort stay below $b$: $S_k\le bk$ at every decision, with $S_k=\sum_{i<k}c(\bar a_i)$ the effort of the executed chunks $\bar a_i$.

\textbf{Certificate.} A predictor $\hat p_{k,j}(a)$ of the robot (known or learned~\citep{moroncelli2026diffusion}) and a forecast $\hat\xi_{k,j}$ of the hazard cover the $T$ steps of the chunk and $H$ further steps of a backup that holds or brakes. A chunk is \emph{certified} if
\begin{equation}\label{eq:cert}
\min_{1\le j\le T+H}\Big[h\big(\hat p_{k,j}(a),\hat\xi_{k,j}\big)-\delta-q_{k,j}-r_{k,j}\Big]\;\ge\;0 .
\end{equation}
The margins $\delta$ and $q_{k,j}$ absorb predictor and forecast errors; with $h_k$ the barrier at the start of the chunk, $r_{k,j}=(1-\alpha)^j\max(h_k,0)$ with $\alpha\in(0,1]$ for $j\le T$ (zero afterwards) is the discrete-time barrier condition~\citep{agrawal2017discrete}. For a budget, a chunk is certified if $S_k+c(a)\le b(k+1)$. We write $\cF_k$ for the set of certified chunks and $\beta_k=\pi(\cF_k\mid o_k)$ for the \emph{feasible fraction}: the probability that one sample of the policy is certified. When we argue with the closed-loop state $s$ we write $\pi(\cdot\mid s)$, $\cF(s)$ and $\beta(s)$. The goal is \emph{joint success}: task completed, request kept at every step.

\section{CAST: Certify, Then Select}\label{sec:method}
\begin{algorithm2e}[t]
\caption{\miss, one decision. Inputs: observation $o$, frozen policy $G$, certificate, value ensemble $Q_1,\dots,Q_J$, number of samples $N$.}\label{alg:cast}
\DontPrintSemicolon
draw $x_0,\dots,x_{N-1}\sim\cN(0,I)$; decode $a_i=G(o,x_i)$ in one batch \tcp*{sample}
$F\gets\{i:a_i\text{ is certified}\}$ \tcp*{certify}
\lIf{$F\neq\emptyset$}{\Return $a_{i^\star}$, $\;i^\star=\arg\max_{i\in F}\frac1J\sum_rQ_r(o,a_i)$ \tcp*[f]{select}}
\lElse{\Return the fallback: an output filter's correction of a sample, or the backup}
\end{algorithm2e}
Algorithm~\ref{alg:cast} is \miss; \miss-V changes only the select line. A correction need not end certified; the backup of the last certified chunk was certified with it (\propref{prop:backupexo}, \tabref{tab:backup}).

\textbf{Certify first.} The certificate is checked before any value is consulted, so the critic never trades safety against task progress and ranks exactly the chunk that is executed. The opposite order, \emph{select-then-filter}, edits the critic's favourite and can execute a chunk the critic never scored.

\textbf{A critic that needs no request.} The ensemble ($J=5$ small networks) regresses the discounted task success of the frozen policy's rollouts, estimating $Q^\pi$, once and before any request is known. The \emph{unanimity guard} variant keeps a certified $a_0$ unless every member prefers another.

\textbf{A critic that knows the request.} When a request changes which behaviour succeeds, as a zone blocking the preferred route does, we condition the critic on it: since the hazard ignores the robot, recorded frozen-policy rollouts can be relabelled with randomly drawn requests (success only if the request is kept from the current step on), with the request parameters as extra inputs.

\textbf{Staying able to act (\miss-V).} A certified chunk can lead to a state from which the frozen policy rarely samples a certified continuation, for example in front of a zone that blocks the route it has started. Define the \emph{viability} $v(s,a)$ of a certified chunk $a$ at state $s$ as the probability that, after $a$, the critic-free certified controller with $m$ samples finds a certified sample at each of the next $L$ decisions, or reaches the goal first. \miss-V estimates it online by the fraction of $M$ model rollouts from the candidate's end state that do so, executes the highest-value candidate with estimate at least $v_{\min}$, and the most viable one if there is none. No critic is retrained; the cost is up to $NMmL$ extra samples per decision ($m$ per continuation step; \tabref{tab:cost}). \emph{Learned viability (\miss-VL)} simulates nothing: its only model is the certificate's one-chunk prediction. Each chunk in logs of the critic-free controller is labelled 1 if the next $L$ decisions all had a certified sample and nothing failed; an ensemble fitted to these labels replaces the rollouts.

\section{Theory: What the Certificate, the Samples and the Critic Each Provide}\label{sec:theory}

\subsection{Safety comes from the certificate, not from the critic}\label{sec:safety}
\begin{assumption}[Prediction errors]\label{ass:err}
On every executed step the robot is within $\delta$ of its prediction and the hazard within $q_{k,j}$ of its forecast: $\|p_{kT+j}-\hat p_{k,j}(\bar a_k)\|\le\delta$ and $\|\xi_{kT+j}-\hat\xi_{k,j}\|\le q_{k,j}$ for $j\le T$. The initial state is safe.
\end{assumption}
The forecast margins can be calibrated once, for a whole episode of at most $K_{\max}$ decisions. Fix a miscoverage level $\varepsilon\in(0,1)$ and scales $\sigma_{k,j}>0$ and score a recorded hazard episode by its largest normalised forecast error, $\Sigma=\max_{k<K_{\max},\,j\le T+H}\|\hat\xi_{k,j}-\xi_{kT+j}\|/\sigma_{k,j}$. From $n\ge1/\varepsilon$ calibration episodes take the $\lceil(n+1)(1-\varepsilon)\rceil$-th smallest score $s_\varepsilon$ and set $q_{k,j}=s_\varepsilon\sigma_{k,j}$.
\begin{theorem}[Safety]\label{thm:safety}
\textbf{(a)} Under Assumption~\ref{ass:err}, the request holds on every step of every certified chunk that is executed, for any number of samples, any critic and any selection rule.
\textbf{(b)} Suppose the initial state is safe, the hazard, its tracker and its forecaster do not depend on the robot, and calibration and deployment episodes are exchangeable. With the margins above, for every controller and jointly over calibration and deployment,
\begin{equation}\label{eq:episode}
\Prob\big[\,h(p_t,\xi_t)<0\text{ for some }t\,\big]\;\le\;\varepsilon+\Prob[\mathcal B],
\end{equation}
where $\mathcal B$ is the event that at some executed step $\|p_t-\hat p_{k,j}\|>h(\hat p_{k,j},\hat\xi_{k,j})-q_{k,j}$; for a certified chunk the threshold $h(\hat p_{k,j},\hat\xi_{k,j})-q_{k,j}$ is at least $\delta$, so $\mathcal B$ requires a prediction error above $\delta$.
\end{theorem}
\emph{In words.} (a) The critic can be arbitrarily wrong without making a certified step unsafe. (b) The hazard ignores the robot, so one calibration of the forecast serves a filter and \miss alike (\tabref{tab:calib}); the robot's own prediction error, $\Prob[\mathcal B]$, is not calibrated. With an exact predictor, the last certified chunk's backup keeps a static request (\propref{prop:recursive}) and, on the calibration event, a moving one through its $H$ steps (\propref{prop:backupexo}).

\subsection{A certified sample stays close to the policy; a projection does not}\label{sec:minimal}
Fix one decision and its state. Write $\cF$ for the certified set, $\beta=\pi(\cF)>0$ for the feasible fraction, and $\pi_\cF=\pi(\cdot\mid\cF)$ for the policy conditioned on being certified. A batch of $N$ independent samples contains a certified one with probability $p_N=1-(1-\beta)^N$, the \emph{availability}.
\begin{theorem}[Distance to the policy]\label{thm:minimal}
\textbf{(i)} Among all action distributions that produce only certified chunks, $\pi_\cF$ is the unique closest to the policy in relative entropy, at distance $\KL(\pi_\cF\|\pi)=\log(1/\beta)$. Executing the first certified sample of a batch, when there is one, draws exactly from $\pi_\cF$.
\textbf{(ii)} Let $\rho$ be the law of the chunk that \miss executes, given that the batch contains a certified sample, with any critic and any rule. Then
\begin{equation}\label{eq:kl}
\log\frac1\beta\;\le\;\KL(\rho\,\|\,\pi)\;\le\;\log\frac1\beta+\log\frac{N\beta}{p_N}\;\le\;\log\frac1\beta+\log N .
\end{equation}
\textbf{(iii)} Suppose $\pi$ has a density and a filter maps a set of actions of positive probability onto a set of zero volume, as a projection onto the boundary of the certified set does. Then, with positive probability, the filter's output lies in a set $Z$ with $\pi(Z)=0$: its law is not absolutely continuous with respect to $\pi$, and its relative entropy to the policy is infinite.
\end{theorem}
\emph{In words.} Enforcing a request costs at least $\log(1/\beta)$; selecting pays at most $\log N$ more, as in best-of-$N$ alignment~\citep{beirami2024bestofn}, and executes a chunk the policy could have produced. A projection leaves the policy's support, however small the edit; how much that matters depends on where the chunk leads (\propref{prop:authority}, \secref{sec:zone}). Our fallbacks fall under (iii).

\subsection{What ranking adds, and what limits it}\label{sec:value}
Compare \miss with the same controller without a critic. Let $\mu_0$ be \emph{critic-free selection} (execute the first certified sample) and $\mu$ be \miss; both use the same $N$, certificate and fallback law on an empty batch. Let $V^\mu$ and $V^0$ denote their expected joint success over an episode of at most $K_{\max}$ decisions, and $Q^0(s,a)$ the value of executing $a$ in state $s$ and following $\mu_0$ afterwards; here $s$ is the closed loop's Markov state (robot, hazard, time, spent budget, memory, violation flag). At a state with batch $a_0,\dots,a_{N-1}$, let $F$ be the indices of the certified samples and $\bar a$ the chunk \miss executes, and define
\begin{align*}
\Omega_N(s)&=\E\big[\max_{i\in F}Q^0(s,a_i)\;\big|\;F\neq\emptyset\big]-\E_{a\sim\pi_\cF}Q^0(s,a) &&\text{(the most any ranking can gain),}\\
\Lambda_N(s)&=\E\big[\max_{i\in F}Q^0(s,a_i)-Q^0(s,\bar a)\;\big|\;F\neq\emptyset\big] &&\text{(what the critic's errors lose),}
\end{align*}
both zero where $\beta(s)=0$.
\begin{theorem}[Closed-loop gain of ranking]\label{thm:selection}
Let $d_k^\mu$ be the distribution of the state that \miss visits at decision $k$. Then
\begin{equation}\label{eq:selection-value}
V^\mu-V^0=\textstyle\sum_{k<K_{\max}}\E_{s\sim d_k^\mu}\big[\,p_N(s)\,\{\Omega_N(s)-\Lambda_N(s)\}\,\big].
\end{equation}
With $\sigma_\cF^2(s)$ the variance of $Q^0(s,a)$ under $\pi_\cF$,
\begin{equation}\label{eq:closed-ceiling}
0\;\le\;p_N\,\Omega_N\;\le\;\sigma_\cF\sqrt{(N\beta-p_N)/2}\;\le\;\sigma_\cF\sqrt{(N-1)/2}.
\end{equation}
Suppose \miss executes the certified sample with the highest score $\hat Q$, and there are an offset $u(s)$ and a set $E$ of chunks with $\pi_\cF(E)\le\eta$ such that $|\hat Q(s,a)-Q_\star(s,a)-u(s)|\le\epsilon_Q$ for every certified $a$ outside $E$, for some target $Q_\star$. Then
\begin{equation}\label{eq:closed-mismatch}
\Lambda_N\;\le\;2\epsilon_Q\;+\;\E\big[\operatorname{osc}_{i\in F}\,(Q^0-Q_\star)(s,a_i)\;\big|\;F\neq\emptyset\big]\;+\;1-(1-\eta)^N,
\end{equation}
where $\operatorname{osc}$ is the largest minus the smallest value. A critic that is exact for $Q^0$ gives $V^\mu\ge V^0$.
\end{theorem}
\emph{In words: three levers.} Ranking helps only if three things meet on the states \miss visits. \emph{Availability} $p_N$: a certified sample must exist. \emph{Spread} $\sigma_\cF$: certified samples must differ in value; what ranking can gain grows at most as $\sqrt N$. \emph{A reliable critic for the right target}: a task critic ($Q_\star=Q^\pi$) suffices unless the request changes which certified samples succeed, as when a zone blocks the policy's preferred route. The last term grows with $N$: keep $N$ small for a weak critic.

\textbf{Availability along the way.} Certification constrains the next chunk only, so availability can be lost later: a certified chunk may lead to states where the policy rarely samples a certified continuation. Write $f$ for the (exact, deterministic) state transition of a chunk. If from every state of a set $\mathcal K$ one sample is certified and leads back into $\mathcal K$ with probability at least $\beta_{\min}$, a controller that executes such samples keeps one available for $\ell$ decisions except with probability $\ell(1-\beta_{\min})^N$ (\propref{prop:viab}, Appendix~\ref{app:proofs}). Viability theory~\citep{aubin1991viability} and reachability~\citep{nakamura2025generalizing} compute viable sets from the dynamics alone; this one depends on the policy (\figref{fig:viability}).

A controller is \emph{selection-only} with $N$ samples if at each decision it executes a certified one of $N$ fresh samples of $\pi(\cdot\mid s)$, and fails if none is certified. Define the \emph{safe control authority} of the frozen policy by $W_{0,N}\equiv1$ and $W_{\ell,N}(s)=\E\max_{i:a_i\in\cF(s)}W_{\ell-1,N}(f(s,a_i))$, the expectation over $a_{0:N-1}\sim\pi(\cdot\mid s)$, with $\max\emptyset=0$ and $W_{\ell,N}\equiv1$ at goal states, which are absorbing.
\begin{proposition}[Finite-sampling generative control authority]\label{prop:authority}
Under the regularity conditions of Appendix~\ref{app:proofs}: (i) $W_{\ell,N}(s)$ is the largest probability, over selection-only controllers with $N$ samples, that each of the first $\ell$ batches contains a certified sample or a goal state is reached first; executing $\arg\max_{i:a_i\in\cF(s_k)}W_{\ell-k-1,N}(f(s_k,a_i))$ at decision $k$ attains it. (ii) $W_{\ell,N}$ is nondecreasing in $N$; $W_{\ell,1}$ is the probability that the policy's own samples are certified for $\ell$ decisions. (iii) As $N\to\infty$, $W_{\ell,N}\uparrow W_{\ell,\infty}\in\{0,1\}$, with $W_{\ell,\infty}(s)=1$ if and only if $\pi(\{a\in\cF(s):W_{\ell-1,\infty}(f(s,a))=1\}\mid s)>0$: viability up to $\pi$-null sets of chunks, and $0$ wherever the certified chunks with $W_{\ell-1,\infty}=1$ form a $\pi$-null set, even if such chunks exist.
\end{proposition}
\emph{In words.} Safe control authority belongs to the dynamics, the request \emph{and} the frozen policy with its $N$ samples: sampling helps up to $\pi$-null sets and no further, and a filter reaches beyond them only by leaving the policy's distribution (\thmref{thm:minimal}). \miss-V's $v(s,a)$ is the success probability of one selection-only controller from $f(s,a)$, so $v(s,a)\le W_{L,m}(f(s,a))$ (estimation error: Proposition~\ref{prop:est}). The Bellman-style recursion follows established planning over random action sets~\citep{boutilier2018stochastic,ghasemipour2021emaq}. Our contribution is its interpretation as the finite-sampling safe-control authority of a frozen generative policy.

\section{Experiments}\label{sec:experiments}
\begin{table}[t]
\centering
\caption{Robomimic Can, joint success under five requests (200 paired episodes each). Methods without a critic stay within 0.08 of the output filter except under R3; \miss adds 0.28--0.36 to it. The R3 filter caps every chunk; with \miss's cap it reaches \RgBudScaleQ{} on \figref{fig:regime}b's scenarios (\miss: \RgBudCastQ).}
\label{tab:main}
{\scriptsize\setlength{\tabcolsep}{3.6pt}\begin{tabular}{lcccccc}
\toprule
Method & R1 moving & R2 fast & R3 effort & R4 ceiling & R5 pillar & max.\ viol. \\
 & obstacle & obstacle & budget & + wall & & (\%) \\
\midrule
Frozen policy & 0.34 & 0.28 & 0.00 & 0.08 & 0.17 & 73.5 \\
Output filter (CBF-QP) & 0.61 & 0.49 & 0.37 & 0.47 & 0.53 & 2.5 \\
In-denoising projection & 0.60 & 0.51 & 0.39 & 0.45 & 0.52 & 1.0 \\
DPCC & 0.58 & 0.57 & 0.51 & 0.40 & 0.53 & 1.5 \\
Latent noise optimisation & 0.59 & 0.43 & 0.26 & 0.42 & 0.47 & 1.5 \\
Critic-free selection & 0.60 & 0.53 & 0.39 & 0.47 & 0.49 & 1.5 \\
\midrule
Select-then-filter & 0.89 & 0.81 & 0.46 & 0.72 & 0.86 & 1.5 \\
\midrule
\miss & 0.89 & 0.85 & \textbf{0.70} & 0.79 & 0.86 & \textbf{0.5} \\
\miss + guard & \textbf{0.90} & \textbf{0.86} & \textbf{0.70} & \textbf{0.80} & \textbf{0.89} & \textbf{0.5} \\
\bottomrule
\end{tabular}
}
\end{table}
\textbf{Setup.} \emph{Systems:} robomimic~\citep{mandlekar2021robomimic} Can with the state-based and camera-only diffusion policies released with DPPO~\citep{ren2024dppo}, and Square with three diffusion policies trained on human demonstrations; a unicycle with known dynamics whose policies pass an obstacle above or below; a planar quadrotor and the MuJoCo humanoid, three policies each. \emph{Requests:} a cylinder sweeping across the workspace; keep-out volumes; a budget on command effort. \emph{Baselines:} the output filter (a CBF-QP on the chunk; scaling for a budget), projection inside the denoising loop~\citep{xiao2023safediffuser}, DPCC~\citep{romer2025dpcc}, noise optimisation~\citep{chen2026inspo}, critic-free selection, select-then-filter, a predictive safety filter~\citep{wabersich2021predictive}, and MPPI~\citep{williams2017information}, which plans with the model, map and goal; all certified methods share the certificate, predictor and margins. \emph{Protocol:} $N=8$; 200 episodes per manipulation cell, 500 per unicycle cell; brackets: paired bootstrap 95\,\% intervals.

\subsection{CAST keeps the task while it enforces the request}\label{sec:can}
\emph{Editing does not help the task; selecting does} (\tabref{tab:main}; rollouts: Figures~\ref{fig:rollouts}, \ref{fig:gallery}). The frozen Can policy ignores the request and violates it in up to 73.5\,\% of episodes. The output filter cuts violations to at most 2.5\,\% and reaches 0.37--0.61 joint success. The other methods without a critic land within 0.08 of it, except under the budget: they change how the constraint is met, not how well the task is done. \miss reaches 0.70--0.89 and violates in at most 0.5\,\% of episodes.

\begin{figure}[t]
\centering
\includegraphics[width=.95\textwidth]{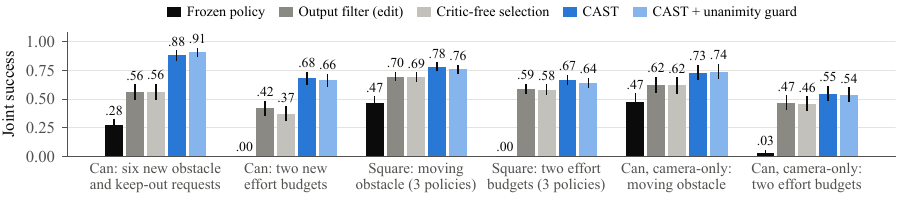}
\caption{\textbf{New requests, frozen models.} Joint success on requests fixed in advance and evaluated once (right: the camera-only Can policy under R1 and two budgets, 0.20 and 0.30). Whiskers: 95\,\% intervals over shared scenarios. Select-then-filter and all other arms: \tabref{tab:fig2cells}.}
\label{fig:headline}
\end{figure}

\emph{The gain holds on new requests and policies}, fixed in advance and scored once (\figref{fig:headline}). Over the filter, \miss gains \HlCanGainPt on six new Can geometries, \HlSqGainPt on Square (positive for each of three policies), \SbCastFilterPt under Square effort budgets, and \TauGapOnePt and \VbCastFilter{} for a camera-only Can policy under R1 and under budgets. Every interval excludes zero (\figref{fig:forest}a).

\emph{Where the gain comes from.} The critic avoids weak samples (best of eight without a request: \CdCanFreeGreedyR{} against \CdCanFreeAR), and shuffling its scores removes the gain (Can \CdCanObstRankPt, \CdCanPillarRankPt; Square \CdSquareObstRankPt).

\emph{Certify first, or rank first?} Select-then-filter has the same samples and critic. For obstacles and keep-out volumes, where its edit is a small detour, it is close to \miss (\HlCanStf{} against \HlCanCast{} on the six new geometries; every setting: \figref{fig:forest}b); there certifying first costs nothing and keeps \thmref{thm:minimal}'s guarantee. Under a budget the edit is large and the order decides: on the two new budgets \miss reaches \HlBudCast{} against \HlBudStf{} for select-then-filter (\BuGreedySTFR) and \HlBudScale{} for scaling. Square's budgets cut its chunks less, and there ranking first does better (\SbCastStf{} for \miss).

\subsection{When does selection help? Moving the three levers}\label{sec:regime}
\begin{figure}[t]
\centering
\includegraphics[width=.9\textwidth]{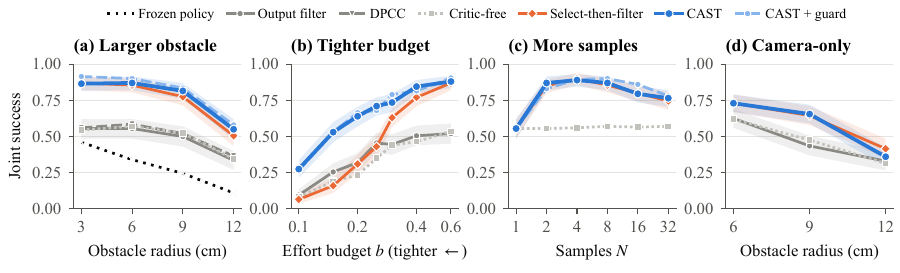}
\caption{\textbf{The levers of \thmref{thm:selection} on Can} (200 shared scenarios per point; bands: 95\,\% intervals). (a) \miss keeps its lead as the obstacle grows. (b) Under a tight budget, ranking first falls to plain scaling. (c) Two to four samples give the whole gain; large batches lose part of it. (d) The camera-only policy: the lead holds at 9\,cm, not at 12\,cm.}
\label{fig:regime}
\end{figure}
We move each lever on Can (\figref{fig:regime}). \emph{A larger obstacle} (a) blocks more of the workspace. \miss keeps \RgFilterLo{} to \RgFilterHi{} over the filter at every size, while DPCC and critic-free selection stay with the filter; select-then-filter is level for small obstacles and behind at 9 and 12\,cm (\RgStfLargePool{} for \miss). \emph{A tighter budget} (b) makes the edit large. As it tightens, select-then-filter falls to plain scaling, whereas \miss executes a chunk that fits whenever one exists: at $b=0.15$, \RgBudCastB{} against \RgBudStfB{} and \RgBudScaleB{}; pooled over $b\le0.30$, \RgBudPoolPt. \emph{More samples} (c): two already give \RgNTwo{} and four \RgNFour, but $N=32$ falls back to \RgNThirtyTwo{} (\RgNDropPt from $N=8$): the critic's own errors are selected, as the last term of~\eqref{eq:closed-mismatch} allows. \emph{The camera-only policy} (d) gains most where the edit costs most (\VrNinePt at 9\,cm), but its certified chunks are nearly interchangeable: under the state-based critic the best of eight is worth \SpCam{} more than their mean (\SpState{} for the state policy), so even that critic, reading the true object state, picks no better (\OrCam). Spread, not the critic, limits the gain. \emph{When a lever is weak} the gain vanishes: on robomimic Lift every certified controller reaches 0.94--0.97, and with a pillar in Square's insertion path \CdSquarePillarFallback{}\,\% of batches are empty and random scores do as well (\CdSquarePillarRankPt).

\subsection{Blocked routes: a critic that knows the request, or viability}\label{sec:zone}
When a request blocks the route a policy has committed to, one-step certification stalls, and looking ahead at the policy's own samples fixes it without retraining. On the unicycle, a keep-out zone blocks one of two routes. The task critic cannot know it; once committed, the only certified chunks brake, and the robot stops short of the goal in \ZfStallFilter{}\,\% of episodes with the filter and \ZfStallTask{}\,\% with \miss, while select-then-filter, whose edit slides along the zone, reaches \ZfStfTask{} against \ZfTask. A predictive safety filter with a 20-step horizon stalls too (joint \ZfPsf). Two remedies work on new zones of the family (\figref{fig:concept}d). \emph{A critic that knows the request}, trained once on zones drawn from a family, lifts \miss to \ZfCast, \ZfTargetGainPt over a matched task critic; with it, select-then-filter does as well (\ZfStfReq). \emph{Viability} needs no retraining at all: \miss-V, with the task critic and $L=8$, reaches \CvEight{} and stalls in \CvStallEight{}\,\% of episodes, beating \miss by \CvTaskPt (\figref{fig:forest}c) and select-then-filter with the same critic by \CvStfPt. Viability alone does as well (\CvVonly): future sample availability, not task value, decides. The lookahead matters ($L=1,2,4$: \CvOne, \CvTwo, \CvFour), $M$ and $v_{\min}$ little (\tabref{tab:castvabl}); more samples do not replace it and select the critic's errors (\miss with $N=1024$: \CmUni; \tabref{tab:cost}). \emph{Learned from logs} (\miss-VL), viability reaches \MfUni{} (\MfUniGainPt over \miss), stays within 0.03 of model-based \miss-V when the actuators are 15--30\,\% weaker than modelled, and ranks never-executed candidates like the simulation (AUC \CtlAuc{} on the scored runs): future safe-control availability is learnable from interaction data (maps: \figref{fig:concept}c, \figref{fig:boundary}). A success critic on the same 3000 logged episodes is slightly ahead (\CtlVal; \tabref{tab:castvabl}) but needs more of them: with 100, viability keeps \DeVl{} and the success critic \DeSc{} (\figref{fig:diag}a). Both remove the lookahead, which costs \miss-V 10--50 times \miss's decision time (\tabref{tab:cost}). With a crossing pedestrian \miss-V gains $+0.09$ over \miss and matches MPPI; \miss-VL reaches \PedVl{} (\figref{fig:ped}).

\subsection{Dynamic systems: a quadrotor and a humanoid}\label{sec:quad}
\begin{figure}[t]
\centering
\includegraphics[width=.9\textwidth]{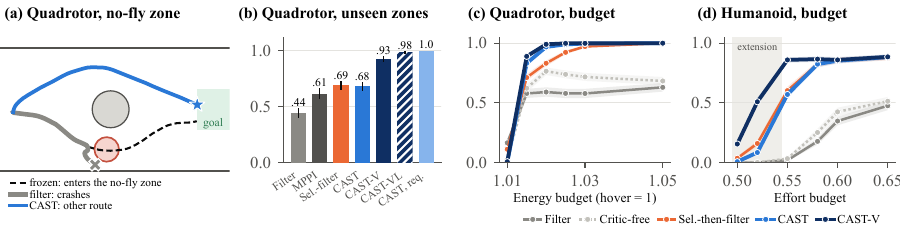}
\caption{\textbf{Dynamic systems, three frozen policies each.} (a) The filter's thrust edits keep a no-fly zone but crash the vehicle; \miss (request-conditioned critic) takes the other route. (b) Unseen zones, 500 scenarios per policy (``req.'': request-conditioned critic). (c) Quadrotor energy budget; at $b=1.01$ almost no chunk fits and scaling does better. (d) Humanoid effort budget, 300 scenarios per policy (shaded: added later). 95\,\% intervals.}
\label{fig:quad}
\end{figure}
A quadrotor couples translation and rotation, so an edited thrust command changes the rest of the flight (\figref{fig:quad}). \emph{No-fly zones.} On zones unseen in training, the output filter never enters the zone but crashes the vehicle in \QzFailFilter{}\,\% of episodes, reaching \QzFilter; MPPI with model, map and goal reaches \QzMppi. \miss with the task critic reaches \QzTask; \miss-V with the same critic and $L=16$ reaches \QvSixteen{} (\QvTaskPt over \miss), and the request-conditioned critic \QzReq. \miss-VL, with no model rollouts, beats model-based \miss-V (\MfQuad). \emph{Energy budget.} The filter scales the thrust down and the vehicle sinks. Pooled over the four budgets 1.015--1.03, \miss beats select-then-filter by \QbStfPt and the filter by \QbFilterPt; \miss-V adds \QbvCastPt. \emph{Humanoid.} Under an effort budget the filter scales the torques and the humanoid falls in \HbFellFilter{}\,\% of episodes. \miss gains \HbFilterPt over the filter over four budgets. Select-then-filter does as well: a small scaling keeps the gait. \miss-V beats both (\HbvCastPt over \miss).

\smallskip\noindent\textbf{Safety.}\label{sec:safetyexp}
With exact models no violation began in a certified chunk; with a learned predictor violations came with the fallback: every violating Square episode of \miss executed a fallback chunk that failed the certificate, and the \SqFbAllN{} whose chunks were all certified had none (\figref{fig:diag}b). \emph{A certified backup aligns the controller with the theorem.} With the pedestrian, executing the last certified chunk's braking backup instead of an uncertified correction, and calibrating the forecast over whole episodes as \thmref{thm:safety}(b) assumes, no contact occurred during a certified step. Contacts did not fall, but all \BkAfterE{} came after the robot had stopped, which is passive safety~\citep{bouraine2012passive}; see \propref{prop:backupexo}. Failures more than halved: \miss rose from \BkCastJEo{} to \BkCastJEb{} (\tabref{tab:backup}).

\section{Conclusion}\label{sec:conclusion}
A frozen diffusion policy can respect a runtime constraint by sampling a few chunks, certifying them, and letting a critic choose: the \textbf{certificate} gives safety, the \textbf{critic} task success. Certifying first matters when a filter's edit is large; viability matters because the physical capacity to act safely and a frozen policy's capacity to sample safe actions differ, and finite sampling makes the difference decisive. \textbf{Limitations.} Selection cannot create behaviour the policy never samples or help a policy that already succeeds; where an edit is benign, ranking first does as well or better; \miss-V costs progress where \miss's choice is already viable. Safety is passive, for hazards that ignore the robot: it covers certified chunks and their backup, not a hazard reaching a robot at rest; tests are simulated.

\smallskip\noindent\textbf{Generative AI use.} Claude (Anthropic) assisted with code, experiments, writing, and literature search; the authors are responsible for the manuscript.

\bibliography{refs}

\clearpage
\appendix
\raggedbottom
\section{Proofs}\label{app:proofs}
\textbf{\thmref{thm:safety}(a).} For $j\le T$, by the Lipschitz property of $h$, Assumption~\ref{ass:err} and~\eqref{eq:cert},
\[h(p_{kT+j},\xi_{kT+j})\;\ge\;h(\hat p_{k,j},\hat\xi_{k,j})-\delta-q_{k,j}\;\ge\;r_{k,j}\;\ge\;0 .\]
Every control step after $t=0$ belongs to one executed chunk, and nothing in the argument involves how the executed chunk $\bar a_k$ was chosen. For the budget, induction on $S_{k+1}=S_k+c(\bar a_k)\le b(k+1)$ keeps the budget; the cap $b(k+1)-S_k$ is at least $b>0$, and for an effort $c$ that is continuous and zero at the zero command, as ours is, a sufficiently scaled chunk satisfies it. \hfill$\square$

\textbf{\thmref{thm:safety}(b).} Write $Z=(\xi_t,\nu_t)_{t\ge0}$ for the hazard trajectory and the tracker noise. By assumption the law of $Z$ is the same under every controller, and the forecasts are fixed functions of $Z$. The score $\Sigma(Z)$ therefore has the same law under every controller. The calibration episodes and the deployment episode are exchangeable, and $n\ge1/\varepsilon$ makes the index $\lceil(n+1)(1-\varepsilon)\rceil$ at most $n$, so $\Prob[\Sigma(Z)\le s_\varepsilon]\ge1-\varepsilon$~\citep{vovk2005algorithmic}, jointly over calibration and deployment. On $\{\Sigma\le s_\varepsilon\}$, for every executed step $t=kT+j$,
\[h(p_t,\xi_t)\;\ge\;h(\hat p_{k,j},\hat\xi_{k,j})-\|p_t-\hat p_{k,j}\|-\|\xi_t-\hat\xi_{k,j}\|\;\ge\;h(\hat p_{k,j},\hat\xi_{k,j})-q_{k,j}-\|p_t-\hat p_{k,j}\|,\]
which is non-negative unless $\mathcal B$ occurs; the initial state is safe. Hence $\Prob[\text{unsafe}]\le\Prob[\Sigma>s_\varepsilon]+\Prob[\mathcal B]$. If the executed chunk is certified, $h(\hat p_{k,j},\hat\xi_{k,j})-q_{k,j}\ge\delta+r_{k,j}$, so $\mathcal B$ needs a prediction error above $\delta$. For a chunk that is not certified the slack can be negative and the bound is void. \hfill$\square$

\begin{proposition}[Recursive feasibility]\label{prop:recursive}
Let the request be static and the predictor exact ($\delta=q=0$), and let the backup be a fixed feedback law that reaches rest within $H$ steps and is safe from the initial state. If the controller executes certified chunks, and the backup when there is none, the request holds at every step and the backup stays safe at every decision.
\end{proposition}
\textbf{\propref{prop:recursive}.} Induction hypothesis at decision $k$: the backup path from $s_{kT}$ keeps $h\ge0$ (true at $k=0$ by assumption). Case (i): a certified chunk is executed. The predictor is exact, so its $T$ steps and the $H$ backup steps after it keep $h\ge0$, and the backup is at rest at the end, where a static $h$ no longer changes. Case (ii): no chunk is certified and the backup is executed; its path is safe by the hypothesis. In both cases the backup path from $s_{(k+1)T}$ is the remainder of a safe path that ends at rest, so the hypothesis holds at $k+1$. The statement does not cover a controller that executes an uncertified correction instead of the backup. \hfill$\square$

\begin{proposition}[Backup against a moving hazard]\label{prop:backupexo}
Under the assumptions and margins of \thmref{thm:safety}(b), let the robot predictor be exact ($\delta=0$), the robot start at rest in a safe state, and the backup an open-loop law that every certificate~\eqref{eq:cert} checks for the $H$ steps after its chunk and that brings the robot to rest within $H$ steps. Suppose the controller executes a certified chunk whenever there is one and otherwise the next $T$ steps of the backup that followed the last certified chunk. Then, on the calibration event $\{\Sigma\le s_\varepsilon\}$ of probability at least $1-\varepsilon$, the request holds at every step of every certified chunk and at each backup step executed within $H$ steps after it: a violation can occur only before the first certified chunk or after $H$ consecutive backup steps, with the robot at rest in both cases (passive motion safety in the sense of \citet{bouraine2012passive}).
\end{proposition}
\textbf{\propref{prop:backupexo}.} Fix a decision $k$ at which a certified chunk is executed. On the event, $\|\xi_{kT+j}-\hat\xi_{k,j}\|\le q_{k,j}$ for every $j\le T+H$, because $\Sigma$ is a maximum over all $k<K_{\max}$ and $j\le T+H$. The certificate gives $h(\hat p_{k,j},\hat\xi_{k,j})-q_{k,j}\ge r_{k,j}\ge0$ for $j\le T+H$, where $\hat p_{k,j}$ for $j>T$ is the predicted backup. If the following decisions have no certified chunk (sample or correction), the controller executes the continuation of that same backup, so with an exact predictor the executed positions are $p_{kT+j}=\hat p_{k,j}$ for $j\le T+H$, and by the Lipschitz property $h(p_{kT+j},\xi_{kT+j})\ge h(\hat p_{k,j},\hat\xi_{k,j})-q_{k,j}\ge0$. A new certified chunk restarts the argument at its own decision. Every step is therefore covered until $H$ backup steps have been executed in a row, after which the robot is at rest and no certificate makes a claim about a hazard that then reaches it. \hfill$\square$

\textbf{\thmref{thm:minimal}.} (i) For $\rho\ll\pi$ with $\rho(\cF)=1$, $d\pi_\cF/d\pi=\ind_\cF/\beta$, so
\begin{equation}\label{eq:pyth}
\KL(\rho\|\pi)=\int\log\frac{d\rho}{d\pi_\cF}\,d\rho+\int\log\frac{d\pi_\cF}{d\pi}\,d\rho=\KL(\rho\|\pi_\cF)+\log\frac1\beta ,
\end{equation}
which is minimised uniquely at $\rho=\pi_\cF$; if $\rho\not\ll\pi$ the divergence is infinite. The samples are independent draws from $\pi$ at the fixed state, so the first certified sample in draw order has law $\pi_\cF$. (ii) The lower bound is (i), since $\rho(\cF)=1$. For the upper bound, the executed chunk is one of the certified samples, so for every measurable $E$,
\[\rho(E)=\frac{\Prob[\bar a\in E,\;F\ne\emptyset]}{p_N}\;\le\;\frac{\Prob\big[\bigcup_i\{a_i\in E\cap\cF\}\big]}{p_N}\;\le\;\frac{N\,\pi(E\cap\cF)}{p_N}=\frac{N\beta}{p_N}\,\pi_\cF(E).\]
Hence $d\rho/d\pi_\cF\le N\beta/p_N$ and $\KL(\rho\|\pi_\cF)\le\log(N\beta/p_N)$; insert this in~\eqref{eq:pyth}. Finally $p_N=1-(1-\beta)^N\ge\beta$ gives $N\beta/p_N\le N$. (iii) Let $Z$ be the zero-volume image set. $\pi(Z)=0$ because $\pi$ has a density, while the filter's output lies in $Z$ with positive probability; a law that charges a $\pi$-null set has infinite relative entropy to $\pi$. \hfill$\square$

\textbf{\thmref{thm:selection}.} \emph{Identity.} All values use the same reward (joint success) and the same Markov state; terminal states are absorbing and the horizon is $K_{\max}$. Telescoping $V^0(s_{k+1})-V^0(s_k)$ along the trajectories of $\mu$ gives the performance-difference identity~\citep{kakade2002approximately}
\[V^\mu(s_0)-V^0(s_0)=\textstyle\sum_k\E_{s\sim d_k^\mu}\big[\E_{a\sim\mu(\cdot\mid s)}Q^0(s,a)-V^0(s)\big].\]
At one state, couple $\mu_0$ and $\mu$ on the same batch and the same fallback randomness; only their action marginals enter the identity, so the coupling costs nothing. On an empty batch both execute the shared fallback and the difference vanishes. On a non-empty batch let $Q^{\max}=\max_{i\in F}Q^0(s,a_i)$. The critic-free action is the first certified draw, with conditional law $\pi_\cF$. Therefore
\[\E_{a\sim\mu}Q^0(s,a)-V^0(s)=p_N\Big\{\underbrace{\E[Q^{\max}\mid F\ne\emptyset]-\E_{\pi_\cF}Q^0}_{\Omega_N}-\underbrace{\E[Q^{\max}-Q^0(s,\bar a)\mid F\ne\emptyset]}_{\Lambda_N}\Big\},\]
and $\Omega_N\ge0$ because the first certified sample has law $\pi_\cF$.

\emph{Ceiling.} Conditional on the number $n\ge1$ of certified samples, their chunks are independent draws from $\pi_\cF$. Let $\phi$ be the quantile function of $Q^0(s,a)$ under $\pi_\cF$, with mean $\bar Q$ and variance $\sigma_\cF^2$. The maximum of $n$ independent uniforms has density $nu^{n-1}$, so by Cauchy--Schwarz
\begin{align*}
\E\max_{i\le n}Q^0(s,a_i)-\bar Q&=\int_0^1(\phi(u)-\bar Q)(nu^{n-1}-1)\,du\;\le\;\sigma_\cF\Big(\int_0^1(nu^{n-1}-1)^2du\Big)^{1/2}\\
&=\sigma_\cF\frac{n-1}{\sqrt{2n-1}}\;\le\;\sigma_\cF\sqrt{\tfrac{n-1}2},
\end{align*}
the classical bound on the expected maximum~\citep{hartley1954universal,gumbel1954maxima}. Averaging over $K\sim\mathrm{Binomial}(N,\beta)$ with Jensen's inequality,
\[p_N\Omega_N\;\le\;\sigma_\cF\,\E\sqrt{(K-1)_+/2}\;\le\;\sigma_\cF\sqrt{\E(K-1)_+/2}\;=\;\sigma_\cF\sqrt{(N\beta-p_N)/2}.\]
The last expression increases with $\beta$ and equals $\sigma_\cF\sqrt{(N-1)/2}$ at $\beta=1$.

\emph{Ranking loss.} Let $\mathcal G$ be the event that no certified sample of the batch lies in the exceptional set $E$; given $K=n\ge1$ certified samples, which are independent draws from $\pi_\cF$, $\Prob[\mathcal G\mid K=n]=(1-\eta)^n\ge(1-\eta)^N$. On $\mathcal G$, let $i_0$ maximise $Q^0$ and let $i^\star$ maximise $\hat Q$ among the certified samples. Then
\[Q^0(s,a_{i_0})-Q^0(s,a_{i^\star})=\underbrace{Q_\star(s,a_{i_0})-Q_\star(s,a_{i^\star})}_{\le2\epsilon_Q}+\underbrace{(Q^0-Q_\star)(s,a_{i_0})-(Q^0-Q_\star)(s,a_{i^\star})}_{\le\operatorname{osc}_{i\in F}(Q^0-Q_\star)(s,a_i)} ,\]
where the first bound uses $\hat Q(s,a_{i^\star})\ge\hat Q(s,a_{i_0})$ and $|\hat Q-Q_\star-u(s)|\le\epsilon_Q$ (the offset cancels). Off $\mathcal G$ the loss is at most one, because values are success probabilities. Taking the conditional expectation gives~\eqref{eq:closed-mismatch}. With $Q_\star=Q^0$ the oscillation vanishes, and with $\Omega_N\ge0$ the identity~\eqref{eq:selection-value} gives $V^\mu-V^0\ge-\sum_k\E\big[p_N\{2\epsilon_Q+1-(1-\eta)^N\}\big]$. The bound is for the maximiser of one score; it does not cover the unanimity guard. \hfill$\square$

\begin{proposition}[Generative viability]\label{prop:viab}
Let $\beta_{\mathcal K}(s)=\pi\big(\{a\in\cF(s):f(s,a)\in\mathcal K\}\mid s\big)$ for a set $\mathcal K$ of states, with $f$ exact and deterministic and batches drawn afresh at each decision. Suppose $s_0\in\mathcal K$ and $\beta_{\mathcal K}(s)\ge\beta_{\min}>0$ for every $s\in\mathcal K$. If a controller executes, whenever its batch contains one, a certified sample that leads into $\mathcal K$, then the probability that some batch among the first $\ell$ decisions contains no such sample is at most $\ell(1-\beta_{\min})^N$; on the complementary event every executed chunk is certified and the state stays in $\mathcal K$.
\end{proposition}
\textbf{Proposition~\ref{prop:viab}.} Let $A_k$ be the event that the batch at decision $k$ contains no certified sample leading into $\mathcal K$. On $\bigcap_{i<k}A_i^c$ every executed chunk was certified and led into $\mathcal K$, so $s_k\in\mathcal K$. Given $s_k\in\mathcal K$, the $N$ samples are independent draws from $\pi(\cdot\mid s_k)$, each certified and leading into $\mathcal K$ with probability $\beta_{\mathcal K}(s_k)\ge\beta_{\min}$, so $\Prob[A_k\mid s_k]\le(1-\beta_{\min})^N$. Hence $\Prob[\bigcup_{k<\ell}A_k]\le\sum_{k<\ell}\Prob[A_k\cap\bigcap_{i<k}A_i^c]\le\ell(1-\beta_{\min})^N$. \hfill$\square$

\textbf{Proposition~\ref{prop:authority}.} Hypotheses: standard Borel state and action spaces, $\pi$ a stochastic kernel, $f$ deterministic and measurable, $\cF(s)$ measurable in $(s,a)$; batches are drawn afresh at each decision, independently of the past given the current state; goal states are absorbing. With a stochastic exogenous hazard, (i) and (ii) hold with $W_{\ell-1,N}(f(s,a_i))$ replaced by its expectation over the hazard's transition; (iii) needs the deterministic $f$. A selection-only controller may use the whole history and its own randomisation, independent of future batches given the current state. (i) By induction on $\ell$; $\ell=0$ is trivial. Fix any selection-only controller and condition on the history and on the first batch $a_{0:N-1}$. If no sample is certified the controller has failed. Otherwise it executes some certified $a_j$, possibly at random, and from $s'=f(s,a_j)$ it continues as a selection-only controller whose history is extra randomisation independent of the future batches; by the induction hypothesis it keeps a certified sample in each of the next $\ell-1$ batches, or reaches a goal state first, with probability at most $W_{\ell-1,N}(s')\le\max_{i:a_i\in\cF(s)}W_{\ell-1,N}(f(s,a_i))$. Taking expectations over the batch gives at most $W_{\ell,N}(s)$. The greedy rule, over certified samples with ties to the smallest index, attains each inequality, so it attains $W_{\ell,N}(s)$ (the $\arg\max$ over finitely many samples is measurable). (ii) By induction, $W_{\ell-1,N+1}\ge W_{\ell-1,N}$, and a maximum over $N+1$ samples dominates the maximum over the first $N$, which are $N$ independent samples; so $W_{\ell,N+1}\ge W_{\ell,N}$. With $N=1$ the only choice is the single sample, executed if certified. (iii) Write $g_{\ell,N}(a)=\mathbf 1\{a\in\cF(s)\}W_{\ell-1,N}(f(s,a))$ and $g_\ell$ for the same with $W_{\ell-1,\infty}$. For a bounded measurable $g\ge0$, $\max_{i\le N}g(a_i)\uparrow\operatorname{ess\,sup}_\pi g$ almost surely, hence in expectation. Assume, inductively, $W_{\ell-1,N}\uparrow W_{\ell-1,\infty}$ pointwise. Upper bound: $W_{\ell,N}(s)\le\operatorname{ess\,sup}_\pi g_{\ell,N}\le\operatorname{ess\,sup}_\pi g_\ell$. Lower bound: for $N\ge N_0$, $W_{\ell,N}(s)\ge\mathbb E[\max_{i\le N}g_{\ell,N_0}(a_i)]\to\operatorname{ess\,sup}_\pi g_{\ell,N_0}$, and $\operatorname{ess\,sup}_\pi g_{\ell,N_0}\uparrow\operatorname{ess\,sup}_\pi g_\ell$ because $g_{\ell,N_0}\uparrow g_\ell$ pointwise (for $\theta>0$ the set $\{g_\ell>\operatorname{ess\,sup}g_\ell-\theta\}$ has positive mass, and by continuity of measure so does $\{g_{\ell,N_0}>\operatorname{ess\,sup}g_\ell-2\theta\}$ for large $N_0$). So $W_{\ell,\infty}(s)=\operatorname{ess\,sup}_\pi g_\ell$, which is $\{0,1\}$-valued by induction and equals $1$ exactly when $\pi(\{a\in\cF(s):W_{\ell-1,\infty}(f(s,a))=1\}\mid s)>0$. The classical indicator $V_\ell(s)=1$ if and only if some admissible $a\in\cF(s)$ has $V_{\ell-1}(f(s,a))=1$; a set of positive $\pi$-mass is nonempty, so by induction $W_{\ell,\infty}\le V_\ell$, with strict inequality exactly where $V_\ell(s)=1$ but the certified chunks with $W_{\ell-1,\infty}=1$ form a $\pi$-null set. The same induction shows that a controller whose action law is absolutely continuous with respect to $\pi(\cdot\mid s)$ at every state survives $\ell$ decisions with probability $0$ from any $s$ with $W_{\ell,\infty}(s)=0$. Finally, \miss-V's critic-free controller is selection-only with $m$ samples, so (i) gives $v(s,a)\le W_{L,m}(f(s,a))$. \hfill$\square$

\textbf{Estimated viability.} With $f$ exact and deterministic, let $v(s,a)$ be the probability that the critic-free certified controller ($m$ samples per decision, same certificate) from $f(s,a)$ finds a certified sample at each of the next $L$ decisions, or reaches the goal first, without task failure, and $\gamma(s)=\pi(\{a\in\cF(s):v(s,a)\ge v_{\rm hi}\}\mid s)$.
\begin{proposition}[Estimated viability]\label{prop:est}
With $M$ continuations on fresh latents and threshold $v_{\min}\in(v_{\rm lo},v_{\rm hi}]$, \miss-V executes a certified chunk with $v\ge v_{\rm lo}$ except with probability at most $(1-\gamma(s))^N+\varepsilon_M$, with $\varepsilon_M$ (two binomial tails, below) $\le Ne^{-2M(v_{\min}-v_{\rm lo})^2}+e^{-2M(v_{\rm hi}-v_{\min})^2}$, after which its next batch is empty with probability at most $(1-v_{\rm lo})^{N/m}$. If $\gamma(s_k)\ge\gamma_{\min}$ at every decision before the first failure, an empty batch or a chunk with $v<v_{\rm lo}$ occurs within $\ell$ decisions with probability at most $\ell[(1-\gamma_{\min})^N+\varepsilon_M+(1-v_{\rm lo})^{N/m}]$.
\end{proposition}
\emph{In words.} Policy, estimate and lookahead width each add a term. Our $M=4$ leaves $\varepsilon_M$ loose (\EstEpsB{} for $v_{\rm lo},v_{\rm hi}=0.1,0.9$), but \miss-V executed a chunk of viability below 0.1 when one above 0.9 existed at \EstErrA{} of decisions (below). The episode bound is vacuous at our $N/m=2$ ($(1-v_{\rm lo})^{N/m}=0.81$); it shows how the risk falls with $N/m$ and $M$.

\textbf{Proposition~\ref{prop:est}.} Hypotheses: $f$ is exact and deterministic, each continuation draws fresh latents, and continuations use the same certificate. Exactly, $\varepsilon_M=N\,\Prob[\mathrm{Bin}(M,v_{\rm lo})\ge\kappa]+\Prob[\mathrm{Bin}(M,v_{\rm hi})<\kappa]$ with $\kappa=\lceil v_{\min}M\rceil$. Fix the batch $a_0,\dots,a_{N-1}$ at $s$. With an exact model, the $M$ continuations of a candidate $a$ use fresh latents, so its estimate $\hat v(a)$ is the mean of $M$ independent Bernoulli$(v(s,a))$ variables given the batch; \miss-V accepts $a$ if $\hat v(a)\ge v_{\min}$, i.e.\ if at least $\kappa$ continuations are viable, and executes the highest-ranked accepted certified candidate (lazy evaluation in critic order stops at exactly this candidate), or the most viable one if none is accepted. \emph{Selection.} Let $B$ be the event that no batch member is certified with $v\ge v_{\rm hi}$; $\Prob[B]=(1-\gamma(s))^N$. On $B^c$ pick such a member $g$. Its estimate falls below $v_{\min}$ with probability $\Prob[\mathrm{Bin}(M,v(s,g))<\kappa]\le\Prob[\mathrm{Bin}(M,v_{\rm hi})<\kappa]$, since the binomial upper tail increases with the success probability. Otherwise the accepted set is non-empty and the executed chunk is accepted; it has $v<v_{\rm lo}$ only if some candidate with $v<v_{\rm lo}$ is accepted, which by a union bound over at most $N$ candidates has probability at most $N\,\Prob[\mathrm{Bin}(M,v_{\rm lo})\ge\kappa]$. Hoeffding's inequality bounds both tails by the exponentials. \emph{Transfer.} Let $s'=f(s,a)$ and $\beta(s')=\pi(\cF(s')\mid s')$. If $s'$ ends the task there is no next batch. Otherwise a continuation is viable only if its first batch of $m$ samples contains a certified one, so $v(s,a)\le1-(1-\beta(s'))^m$. \miss-V's next batch consists of $N$ independent draws at $s'$, so it is empty with probability $(1-\beta(s'))^N=\big((1-\beta(s'))^m\big)^{N/m}\le(1-v(s,a))^{N/m}\le(1-v_{\rm lo})^{N/m}$. \emph{Horizon.} At decision $k$ let $D_k$ be the event that the batch has no certified member with $v\ge v_{\rm hi}$, or the executed chunk has $v<v_{\rm lo}$, or the next batch is empty after a chunk with $v\ge v_{\rm lo}$. Conditionally on the past and $s_k$, $\Prob[D_k\mid s_k]\le(1-\gamma(s_k))^N+\varepsilon_M+(1-v_{\rm lo})^{N/m}$; the first batch is empty only on $D_0$'s first part, and every later empty batch or low-viability chunk lies in some $D_k$. A union bound over $k<\ell$ with $\gamma(s_k)\ge\gamma_{\min}$ gives the claim. Nothing is assumed about the critic. \hfill$\square$

\emph{Measured} (unicycle, three policies $\times$ 500 scored scenarios, \miss-V with $M=4$; reference viability of every certified candidate from 64 further continuations, not used to act; its decisions are identical to the scored cells). \EstTable

\textbf{An exact critic with the wrong target can fail.} Consider two certified chunks that commit to different routes; the policy samples route A with probability $p\in(0,1)$ and route B otherwise. On route A the continuation completes the task safely with probability $1-\zeta$. On route B the frozen policy completes the task with probability one, but only by crossing a keep-out zone later; no certified continuation completes route B. The task values are $Q^\pi(A)=1-\zeta$ and $Q^\pi(B)=1$, whereas the values under the request are $Q^0(A)=1-\zeta$ and $Q^0(B)=0$. An exact task critic selects B whenever a B sample appears, so $V^\mu=p^N(1-\zeta)$, below the critic-free value $V^0=p(1-\zeta)$ for every $N>1$. Both first chunks pass the certificate; the failure is in their continuation. On a batch that contains both routes the oscillation term of~\eqref{eq:closed-mismatch} equals one, and the identity~\eqref{eq:selection-value} returns exactly $V^\mu-V^0=(p^N-p)(1-\zeta)$. A critic trained on success under the request ranks A first and attains $V^\mu=(1-(1-p)^N)(1-\zeta)\ge V^0$.

\section{Experimental details}\label{app:details}
\begin{figure}[t]
\centering
\includegraphics[width=\textwidth]{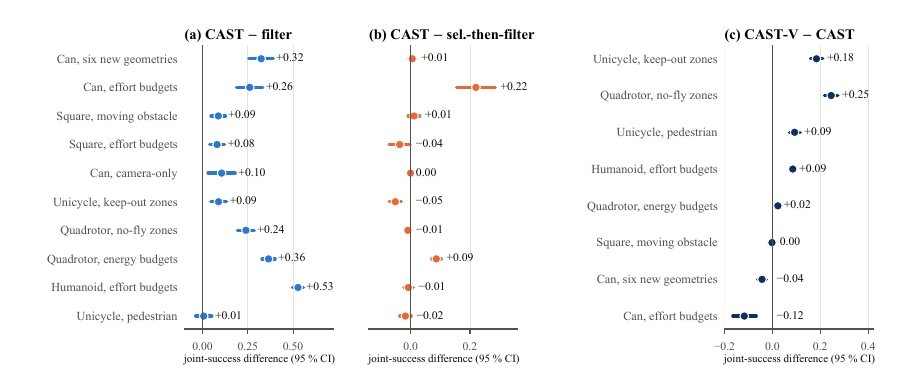}
\caption{\textbf{Every setting at a glance.} Paired differences in joint success with 95\,\% bootstrap intervals over shared scenarios. (a) \miss minus the output filter (for budgets: scaling the policy's own chunk), task critic unless the setting has only a request-conditioned one. (b) \miss minus select-then-filter with the same samples and critic (pedestrian: the request-conditioned pair): certifying first adds where the filter's edit is large (the budgets on Can and on the quadrotor) and is level or slightly behind elsewhere (small detours; mild scalings on Square and the humanoid). (c) \miss-V minus \miss with the same critic: large where a certified step can lead into a state from which the policy cannot continue (a blocked route, a crossing pedestrian, a fall under the humanoid's budget), small under the quadrotor's energy budget, level on Square and negative on Can's new geometries and budgets, where \miss's own choice is already viable and keeping options open costs progress. Pooled cells as in the main text and \tabref{tab:fig2cells}.}
\label{fig:forest}
\end{figure}
\begin{figure}[t]
\centering
\includegraphics[width=\textwidth]{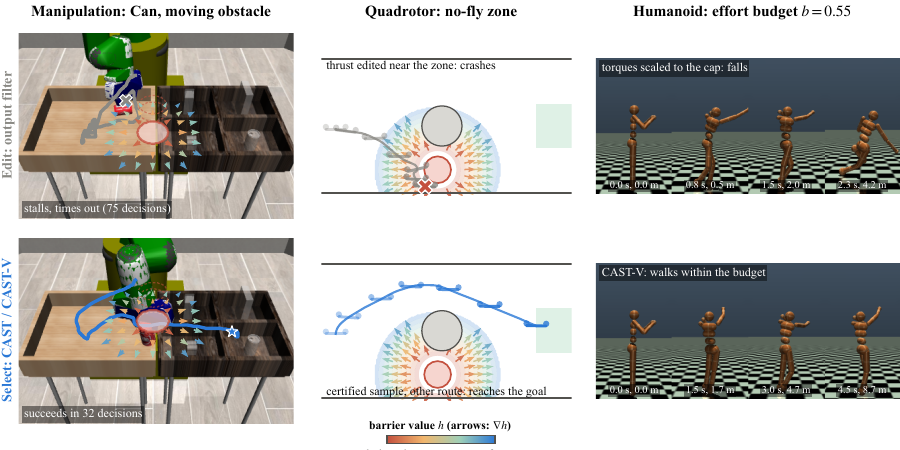}
\caption{\textbf{Rollouts in three simulators: edit (top) or select (bottom).} \emph{Manipulation} (robomimic Can, moving obstacle 0.10\,m / 75, one scored scenario): rendered frames with the hand path, the virtual cylinder at four times (faded) and its barrier-gradient field projected into the camera; the output filter keeps the clearance but stalls until the time limit, \miss delivers the can. \emph{Quadrotor} (no-fly zone unseen in training): the vehicle drawn to scale along its flight over the zone's barrier field; the filter's thrust corrections crash it, \miss (request-conditioned critic) executes a certified sample of the other learned route. \emph{Humanoid} (effort budget $b=0.55$, an illustration scenario): MuJoCo frames; scaling the torques to the cap makes it fall, \miss-V keeps walking within the budget. Arrows: $\nabla h$, coloured by the barrier value $h$ (red: at the boundary; blue: far).}
\label{fig:rollouts}
\end{figure}
\begin{figure}[p]
\centering
\includegraphics[width=.9\textwidth]{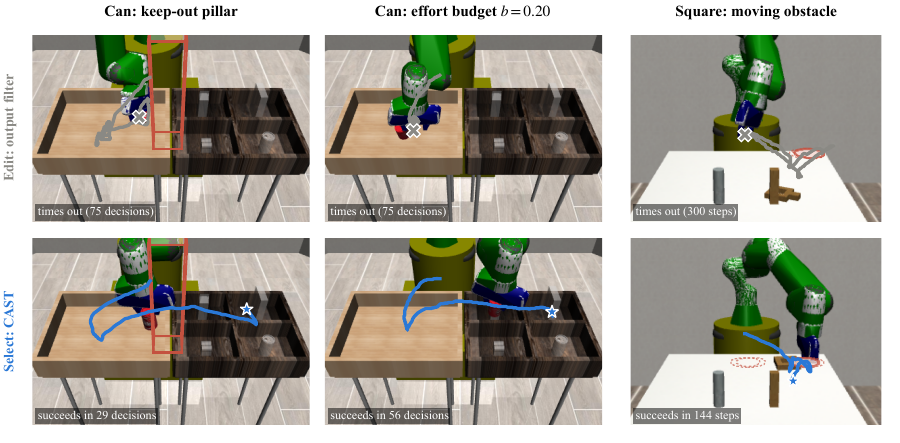}\\[2pt]
\includegraphics[width=.9\textwidth]{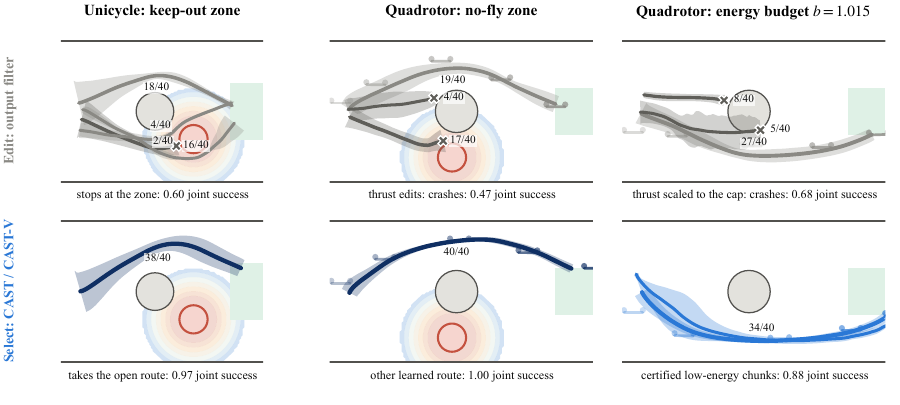}
\caption{\textbf{More rollouts: edit (upper row of each block) or select (lower row).} \emph{Top, manipulation} (rendered; the rendered episodes equal the scored ones). \emph{Can, keep-out pillar} ($+2$\,cm; red wireframe): the filter's corrections keep the hand out of the pillar but it never transports the can; \miss carries it around. \emph{Can, effort budget} $b=0.20$: the filter caps every chunk and the hand stalls with the can; \miss executes certified low-effort chunks and delivers it. \emph{Square, moving obstacle} (policy s0, $-2$\,cm): the filter's fallback holds the nut near the obstacle until the time limit; \miss inserts it. Hand paths projected into the camera. \emph{Bottom, dynamic systems,} drawn to scale from the recorded states: 40 rollouts per panel as mean $\pm$ sd per route and outcome (numbers: rollouts in the group; groups of one are not drawn; crosses: a group that stalls or crashes), one flight with the vehicle drawn. \emph{Unicycle} and \emph{quadrotor no-fly zone}: the zones of Figures~\ref{fig:concept} and~\ref{fig:quad}a fixed, \miss-V with the task critic. \emph{Quadrotor energy budget} $b=1.015$: the first 40 scored scenarios of policy s0, re-simulated for drawing (outcomes differ slightly from the scored run), \miss with the task critic; scaling the thrust to the cap costs altitude.}
\label{fig:gallery}
\end{figure}
\begin{figure}[t]
\centering
\includegraphics[width=\textwidth]{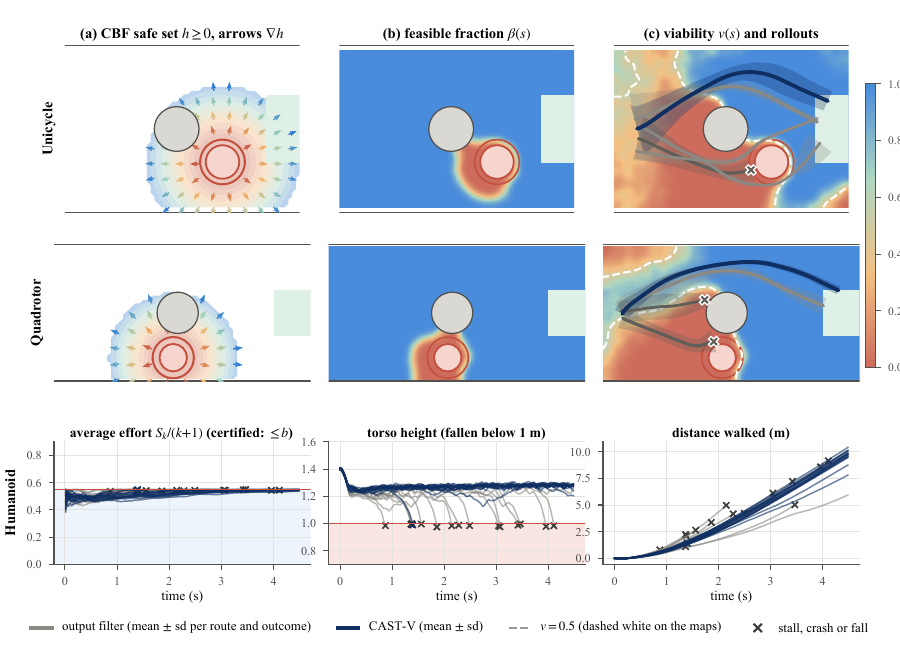}
\caption{\textbf{Safe set, feasible fraction and viability.} Unicycle (Figure~\ref{fig:concept}'s zone) and quadrotor (Figure~\ref{fig:quad}a's zone), on a slice of the state space (heading at the goal and cruise speed; level flight): (a) the CBF safe set $h\geq0$ that a filter protects, with $\nabla h$; (b) the feasible fraction $\beta(s)$, the share of 64 frozen-policy chunks the certificate accepts; (c) \miss-V's viability estimate at the state, $v(s)$: the fraction of 16 critic-free certified continuations started at $s$ that find a certified sample at each of the next $L=8$ decisions (a chunk's viability is $v$ at the state it leads to), its region $v\geq0.5$ (dashed) and 40 rollouts per controller with this zone fixed, as mean $\pm$ sd per route and outcome (filter: grey; \miss-V: dark blue; crosses: a group that stalls or crashes). States inside the safe set and with certified samples available can still be dead ends for this policy; the filter's rollouts end there, \miss-V's avoid them. Humanoid (effort budget 0.55; 16 scenarios): the certified region of the budget, torso height with the fall threshold, and distance walked; the filter rides the cap and falls in 15 of 16, \miss-V in 1. Illustration data, not scored.}
\label{fig:viability}
\end{figure}
\begin{figure}[t]
\centering
\includegraphics[width=\textwidth]{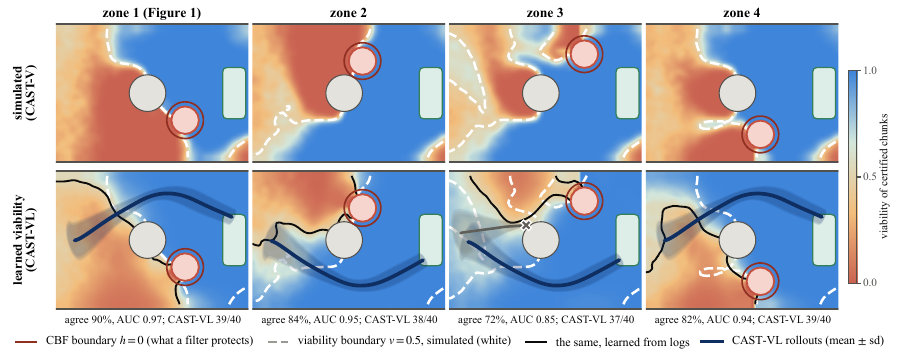}
\caption{\textbf{The boundary a frozen policy needs, simulated and learned.} Four keep-out zones that the learned network never saw (unicycle, policy s0; the slice of the state space of \figref{fig:viability}). Colour: mean viability of the policy's certified chunks at each state (16 samples). The CBF boundary $h=0$ (red) is what a filter protects; the viability boundary $v=0.5$ lies far outside it and moves with the zone. Top: by simulation, as \miss-V estimates it. Bottom: from the network trained on logs of the critic-free controller (\miss-VL), with both boundaries. The two agree on 72--90\,\% of about 42\,000 certified chunks per zone (AUC 0.85--0.97); the learned map is softer inside the dead-end regions, which the logs rarely visit, and matches where the rollouts pass: \miss-VL reaches the goal in 37--39 of 40 rollouts. Illustration data, not scored.}
\label{fig:boundary}
\end{figure}
\begin{figure}[t]
\centering
\includegraphics[width=\textwidth]{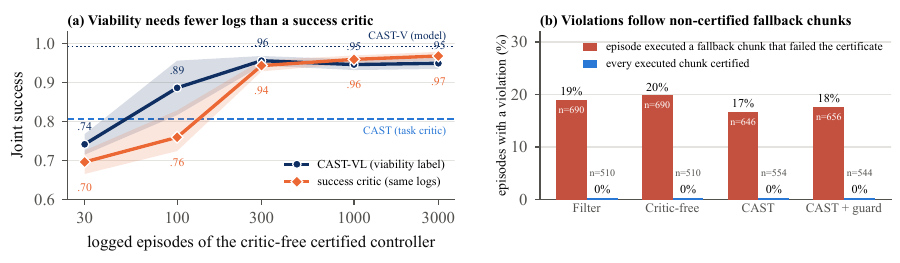}
\caption{\textbf{Two diagnostics.} (a) \emph{Viability or success as the learning target?} Both networks are trained on the same logged episodes of the critic-free controller, with the same features, architecture and recipe; only the label differs (viability of the executed chunk, or the episode's joint success). Unicycle, unseen zones, three policies $\times$ 500 scored scenarios; bands: 95\,\% intervals, and at 100 episodes the range of three disjoint subsets, in each of which viability is ahead. With 300 episodes the two are level (\DeDiffT), with 1000 the success critic is slightly ahead (\DeDiffK), and with 3000 by 0.02. Viability is the cheaper label to learn and needs no task outcome; a success critic can replace it when logs are plentiful. (b) \emph{Where violations come from.} Square, new requests, learned predictor (\SqFbN{} scored episodes per controller): share of violating episodes among those that executed at least one fallback chunk failing the certificate (an uncertified correction, or the hold backup, which is accepted without margins) and among those whose chunks were all certified. \thmref{thm:safety} covers the second group; none of its episodes violated.}
\label{fig:diag}
\end{figure}
\begin{figure}[t]
\centering
\includegraphics[width=\textwidth]{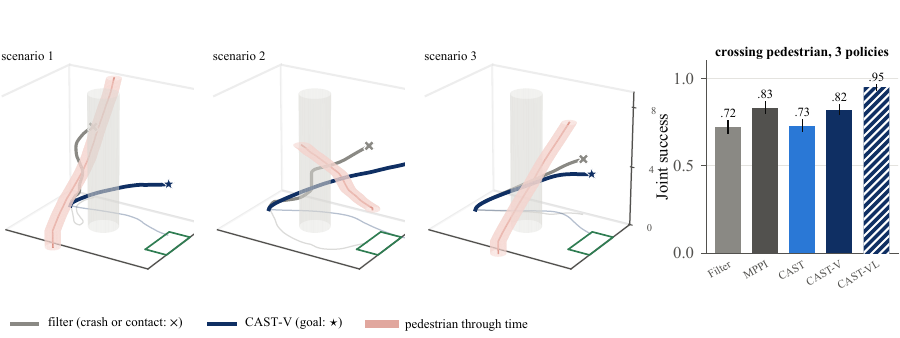}
\caption{\textbf{A moving obstacle in space and time.} Unicycle with a pedestrian crossing the corridor (random-walk speed 0.3--0.6\,m/s, noisy tracker, constant-velocity forecast with per-step conformal margins at level 0.99). Left: three scored scenarios (policy s0) where the filter fails and \miss-V succeeds; the robot's plane is the floor, time goes up, the pedestrian is a slanted tube and the central obstacle a vertical cylinder. The filter's evasive corrections push the robot into the wall (scenarios 1 and 3) or into contact with the pedestrian (scenario 2); \miss-V passes ahead of the pedestrian. Right: joint success over three policies $\times$ 500 shared scenarios. Critics trained on 6000 frozen-policy episodes per policy (task critic held-out $R^2$ 0.19--0.20; request-conditioned critic, which also sees the forecast, 0.68--0.71). \miss-VL: \miss-V's rule with viability learned from 6000 logged episodes per policy of the critic-free certified controller (no crash or contact counts as viable; planned, scored once): \PedVl, \PedVlFilter{} over the filter, \PedVlMppi{} over MPPI and \PedVlCastv{} over model-based \miss-V. A success critic trained on the same kind of logs lifts \miss itself to \PedLogCrit. Planned contrasts: \miss (request critic) $-$ filter $+0.024$ $[-0.018, +0.066]$, so with critics trained on frozen-policy episodes \miss alone does not beat the filter; \miss-V (task critic) $-$ \miss (task critic) $+0.093$ $[+0.071, +0.115]$; \miss-V $-$ MPPI $-0.011$ $[-0.051, +0.028]$. Also: \miss-V $-$ filter $+0.099$ $[+0.059, +0.139]$, $-$ select-then-filter (request critic) $+0.057$ $[+0.033, +0.082]$. Pedestrian contact: 0.9--2.4\,\% for the certified controllers with frozen-policy critics (tracker noise), 0.1\,\% for \miss-VL and the log-trained critic, 0 for MPPI.}
\label{fig:ped}
\end{figure}
\textbf{Robomimic Can.} Policy: the diffusion policy released with DPPO~\citep{ren2024dppo} (MLP denoiser, 20 DDIM steps, chunks of four 7-dimensional commands at 20\,Hz, at most 300 control steps). It completes the task in \CdCanFreeA{} of episodes without any request. Critic: five MLPs on (state, chunk) regressing Monte Carlo task success with discount 0.99 on 2000 rollouts of the frozen policy. Hand predictor: linear in the chunk's cumulative commands and the previous chunk, fitted by least squares on 2000 episodes; $\delta=2$\,cm. Requests of \tabref{tab:main}: R1, a vertical cylinder of radius 6\,cm that sweeps the transport corridor sinusoidally (amplitude 0.15\,m, period 100 steps, random phase); R2, the same cylinder faster than the hand can move (amplitude 0.18\,m, period 50); R3, a budget $b=0.25$ on the mean squared position command; R4, a ceiling and a workspace wall; R5, a non-convex keep-out pillar. The obstacle is observed at the current and previous step and extrapolated at constant velocity; no controller sees its future. For R1, the new obstacle requests and the sweeps the certificate uses $\alpha=1$ and $H=0$: it checks the predicted clearance of the four steps of the chunk, with no rate term and no backup steps. R2 uses $\alpha=0.3$, $H=4$ and conformal forecast margins. When no sample is certified, the obstacle and keep-out controllers apply the output filter (at most four relinearised projections, which need not end certified) to a sample; the budget controllers scale the sample of least effort to the cap. New requests of \figref{fig:headline}: obstacle amplitudes $\{0.10,0.20\}$\,m with periods $\{75,125\}$, the pillar shifted by $\pm2$\,cm, and budgets 0.20 and 0.30. The camera-only policy is DPPO's image policy; its critic reads the policy's own 137-dimensional embedding.

\textbf{Robomimic Square.} Three diffusion policies (seeds of one recipe) trained on the proficient-human demonstrations: 1-D convolutional U-Net, 16 DDIM steps, chunks of sixteen commands of which eight are executed. Each policy has its own critic and hand predictor, fitted on 2000 of its rollouts; $\delta=3$\,cm, the 95th percentile of the predictor's error. Request: a cylinder of radius 6\,cm moving across the path to the peg (amplitude 0.15\,m, period 100), forecast with constant acceleration, at its original position and shifted by $\pm2$\,cm.

\textbf{Unicycle.} State $(x,y,\theta,v)$, time step 0.1\,s, bounded acceleration and turn rate, known exactly to every controller. The robot crosses a corridor ($|y|\le1.5$) around a disc obstacle (centre $(2,0)$, radius 0.4) to a goal region within 120 steps. The diffusion policy (MLP, 10 DDIM steps, chunks of four controls) is trained on 3000 demonstrations that pass above or below the obstacle and succeeds in 0.98 of episodes without a request. Request: a keep-out disc of radius 0.3 on one side. In the family study its centre is drawn per episode from $x\sim U[2.3,3.0]$, $|y|\sim U[0.55,0.95]$. The request-conditioned critic has the architecture of the task critic plus five inputs (zone position relative to the robot, side, current barrier value, time). For each of three policies it is trained on 20\,000 frozen-policy rollouts labelled with zones drawn from the family; the matched task-target critic uses the same rollouts, inputs, initialisation and optimiser, and only the label differs. Evaluation zones are drawn independently. Certified controllers share an 8-step braking backup and $\alpha=0.5$. MPPI (256 samples, horizon 20 steps of the 120) and its hard-constrained variant (same horizon; candidates must pass the certificate) know the dynamics, the map, the goal and the zone, and do not use the policy; they are short-horizon references, not upper bounds. The extra latents of a batch are drawn from a separately seeded generator, so every batch holds $N$ independent samples, as the theory assumes.

\textbf{Planar quadrotor.} State $(x,z,\theta,\dot x,\dot z,\dot\theta)$ at 50\,Hz with collective thrust $F=mg(1+0.8u_0)$ and torque $0.25u_1$; known exactly to every controller. Task: from the left to a goal box within 8\,s, without touching floor, ceiling or a central disc obstacle. Three diffusion policies (MLP, 10 DDIM steps, chunks of eight controls) are trained on 3000 expert demonstrations each, half above and half below the obstacle; task critics (held-out $R^2$ \QCriticRtwo) and request-conditioned critics (held-out AUC \QReqAuc) use 4000 frozen-policy rollouts, each labelled with four zones from the family. Certificate: the chunk and a 25-step brake-to-hover backup, rolled out with the exact model, $\alpha=0.5$; budget: exact prefix inequality on $c=\overline{(F/mg)^2}$. The output filter is the minimal-norm correction (finite-difference linearisation, three relinearisations; for the budget, thrust scaling), shared as the fallback. MPPI: 512 samples, horizon 2\,s, soft request penalties, and the shared certificate on what it executes; without a request it succeeds in 0.69 of development episodes. The budget levels and the planner horizon were chosen on development scenarios disjoint from the scored ones.

\textbf{Humanoid.} Gymnasium Humanoid-v5 (MuJoCo, 17 actuators, 15\,ms control step). Three diffusion policies (MLP 3$\times$512, 10 DDIM steps, single-action chunks, full 348-dimensional observation) are trained on 2M pairs from Farama Minari's humanoid expert-v0 and medium-v0 demonstrations (SAC rollouts), 80k steps on one GPU (75\,s per policy). Success: no fall (the environment's healthy-height termination) within 300 steps and at least 1\,m forward. Critics: five MLPs on (observation, action), discounted success (0.995 per step) from 1500 frozen-policy rollouts per policy (held-out $R^2$ 0.70--0.76). Budget: $c=\|a\|^2/17$ per step on actions normalised to $[-1,1]$, prefix average at most $b$; the filter scales the policy's own action to the cap, the selectors' fallback scales the least-effort sample.

\textbf{CAST-V.} Viability is estimated with $M=4$ continuations of the critic-free certified controller from each certified candidate's end state ($m=4$ policy samples per step, exact model, the request's certificate); a continuation is viable if it finds a certified sample at each of the next $L$ decisions without a task failure, or reaches the goal. $v_{\min}=0.5$; unicycle $L=8$ (3.2\,s), quadrotor $L=16$ (2.6\,s), humanoid $L=20$ steps (0.3\,s; a fall in a continuation is not viable). Candidates are evaluated lazily in critic order (the first with $\hat v\ge v_{\min}$ is executed; same choice as evaluating all). \emph{Robomimic.} The model is a shadow copy of the simulator restored to the current state (MuJoCo state, clock, controller goals and gripper ramp), which a real robot would not have; $L=8$ decisions on Can and $L=4$ on Square (32 steps each); a moving obstacle is forecast at constant velocity from its positions at the real decision, so the exact-model hypothesis of \propref{prop:est} does not hold there. Fidelity of the shadow after one chunk (normalised hand and can position): median 0.003--0.005, maximum 0.04--0.06. With $v_{\min}=0$ the controller reproduced \miss's saved episodes exactly on one 50-episode block of each request and on Square. \miss-V overrode \miss's choice at \RvSwLo--\RvSwHi{}\,\% of decisions on obstacles and Square and \RvSwPbLo--\RvSwPbHi{}\,\% on pillars and budgets; it cost about 35\,s per Can episode and 20\,s per Square episode on one CPU thread (\miss: about 2\,s).

\textbf{Baselines.} \emph{Output filter}: minimal-norm correction of the policy's chunk onto the linearised certificate, solved as an exact dual QP with relinearisation (a CBF-QP on the chunk). For budgets there are two scaling filters: one caps the effort of every chunk at $b$ (\tabref{tab:main}, R3; ``chunk cap'' in \tabref{tab:regime}), a stricter rule than the average budget; the other scales the policy's chunk only as far as the average budget requires, which is \miss with $N=1$ and the like-for-like baseline of Figures~\ref{fig:headline} and~\ref{fig:regime}b. \emph{In-denoising projection}: the same projection applied to the denoised estimate at every DDIM step, in the manner of SafeDiffuser~\citep{xiao2023safediffuser}. \emph{DPCC}~\citep{romer2025dpcc}: our port projects every reverse iterate onto the tightened constraint for $N$ chains started from the latents \miss draws and executes the chain with the smallest cumulative projection. \emph{Noise optimisation}, in the manner of INSPO~\citep{chen2026inspo}: Adam steps on the latent against the predicted violation, through the differentiable sampler. \emph{Critic-free selection}: \miss without the critic. \emph{Select-then-filter}: the sample with the highest value, then the output filter. All arms of a comparison use the same predictor and margins and, where they sample, the same latents; the sampling controllers also share the fallback.

\textbf{Protocol.} Unless stated, $N=8$; 200 episodes per manipulation cell and 500 per unicycle cell; all arms of a comparison run on the same scenarios, with paired bootstrap 95\,\% intervals. The policy, critic and predictor are fixed before the evaluation scenarios are drawn. For four studies the design, scenario seeds and planned comparisons were fixed before any scored episode was run: the new requests (\figref{fig:headline}), the ranking intervention, the Can sweeps (\figref{fig:regime}) and the zone family (\figref{fig:concept}d). The later studies also had their plans fixed before scoring, except the $M$/$v_{\min}$ ablation, the humanoid's two tightest budgets, the cells that complete \tabref{tab:fig2cells} and the violation breakdown of \figref{fig:diag}b. On the new Can and Square requests the planned primary arm was \miss with the guard; plain \miss, reported in the text, was secondary. Not every planned threshold was met: the budget gains on Square and with the camera-only policy were $+0.08$ against a planned $+0.10$, and \miss-V did not improve on \miss in manipulation (\figref{fig:forest}c). Simulations ran in Docker on one workstation (crossing pedestrian and \miss-V ablation: a laptop CPU); a decision of \miss with the guard takes \LatMissCpu{}\,ms on two CPU threads.

\textbf{Additional tables.}\label{app:results} Tables~\ref{tab:fig2cells}--\ref{tab:backup} give the cells of Figure~\ref{fig:headline}, the compute, the \miss-V ablation and the fallback study; Tables~\ref{tab:quad}--\ref{tab:modelfree} give the quadrotor, humanoid, unicycle, Can-sweep and \miss-VL cells; \tabref{tab:calib} gives the calibration check.
\makeatletter\newcommand{\heretable}{\def\@captype{table}}\makeatother

\par\medskip\noindent\begin{minipage}{\textwidth}\heretable\centering
\caption{Every arm in every cell of \figref{fig:headline}, with further camera-only cells (9 and 12\,cm: \figref{fig:regime}d; $\tau$: scale of the policy's initial denoising noise, $\tau=1$ elsewhere) (joint success; 200 shared scenarios per cell; -- not run). Frozen: the policy's own sample, ignoring the request. For budgets, ``Filter'' scales the policy's own chunk to the cap (\miss with $N=1$; the per-chunk cap filter: \tabref{tab:regime}) and select-then-filter selects, then scales.}
\label{tab:fig2cells}
{\scriptsize\setlength{\tabcolsep}{4pt}\begin{tabular}{lccccccc}\toprule
Cell & Frozen & Filter & Crit.-free & Sel.-filter & \miss & + guard & \miss-V\\\midrule
Can, obst. 0.10 m / 75 & 0.27 & 0.58 & 0.58 & 0.89 & 0.91 & 0.92 & 0.89\\
Can, obst. 0.10 m / 125 & 0.28 & 0.56 & 0.54 & 0.89 & 0.89 & 0.90 & 0.90\\
Can, obst. 0.20 m / 75 & 0.38 & 0.56 & 0.56 & 0.83 & 0.85 & 0.90 & 0.83\\
Can, obst. 0.20 m / 125 & 0.36 & 0.58 & 0.56 & 0.88 & 0.90 & 0.92 & 0.88\\
Can, pillar $+2$\,cm & 0.17 & 0.54 & 0.57 & 0.88 & 0.88 & 0.91 & 0.77\\
Can, pillar $-2$\,cm & 0.18 & 0.55 & 0.58 & 0.91 & 0.88 & 0.91 & 0.77\\
\midrule
Can, budget 0.20 & 0.00 & 0.34 & 0.31 & 0.30 & 0.64 & 0.55 & 0.43\\
Can, budget 0.30 & 0.00 & 0.50 & 0.43 & 0.61 & 0.72 & 0.78 & 0.69\\
\midrule
Square, s0, $-2$\,cm & 0.47 & 0.70 & 0.66 & 0.81 & 0.83 & 0.77 & 0.81\\
Square, s0, $+2$\,cm & 0.51 & 0.70 & 0.70 & 0.71 & 0.76 & 0.81 & 0.77\\
Square, s1, $-2$\,cm & 0.44 & 0.68 & 0.69 & 0.76 & 0.74 & 0.72 & 0.75\\
Square, s1, $+2$\,cm & 0.49 & 0.69 & 0.68 & 0.76 & 0.74 & 0.71 & 0.77\\
Square, s2, $-2$\,cm & 0.45 & 0.71 & 0.71 & 0.82 & 0.81 & 0.78 & 0.80\\
Square, s2, $+2$\,cm & 0.46 & 0.70 & 0.70 & 0.78 & 0.81 & 0.79 & 0.80\\
\midrule
Square, budget 0.25 & 0.00 & 0.55 & 0.54 & 0.65 & 0.62 & 0.60 & 0.61\\
Square, budget 0.30 & 0.00 & 0.63 & 0.62 & 0.76 & 0.72 & 0.68 & 0.71\\
\midrule
Camera, R1 & 0.47 & 0.62 & 0.62 & 0.73 & 0.73 & 0.74 & 0.73\\
Camera, obstacle 9\,cm & 0.31 & 0.43 & 0.47 & 0.65 & 0.66 & 0.57 & 0.66\\
Camera, obstacle 12\,cm & 0.14 & 0.33 & 0.32 & 0.41 & 0.36 & 0.37 & 0.40\\
Camera, budget 0.15 & 0.00 & 0.27 & 0.29 & 0.30 & 0.34 & 0.37 & --\\
Camera, budgets 0.20/0.30 & 0.03 & 0.47 & 0.46 & 0.51 & 0.55 & 0.54 & --\\
Camera, $\tau=2$ & 0.42 & 0.55 & 0.60 & 0.66 & 0.72 & 0.66 & 0.78\\
Camera, $\tau=2$, budgets 0.20/0.30 & 0.04 & 0.47 & 0.42 & 0.59 & 0.56 & 0.56 & --\\
\bottomrule\end{tabular}
}
\end{minipage}\par

\par\medskip\noindent\begin{minipage}{\textwidth}\heretable\centering
\caption{Compute per decision on the zone tasks (task critic, mean of three policies). Policy samples and simulated model steps are per decision (\miss-V: worst case $N + NMmL$ samples, certificate rollouts of every sample and $NML$ simulated chunks; lazy search and early stops use fewer); ms: measured, one CPU thread, 500 episodes batched. \miss with $N=256$ and $N=1024$ gives the plain selector about as many samples as \miss-V's unicycle worst case; \miss-VL adds one network call to \miss.}
\label{tab:cost}
{\scriptsize\setlength{\tabcolsep}{5pt}\begin{tabular}{llrrrc}\toprule
System & Controller & Policy samples & Model steps & ms & Joint success\\\midrule
Unicycle & \miss, $N=8$ & 8 & 96 & 1.7 & 0.81\\
 & Select-then-filter & 8 & 12 & 1.5 & 0.86\\
 & MPPI (256 samples, horizon 20) & 0 & 5\,120 & 0.9 & 0.91\\
 & \miss, $N=256$ & 256 & 3\,072 & 33.0 & 0.37\\
 & \miss, $N=1024$ & 1024 & 12\,288 & 146.4 & 0.14\\
 & \miss-V, $L=8$ & $\le$1\,032 & $\le$13\,408 & 18.4 & \textbf{0.99}\\
 & \miss-VL & 8 & 96 & 1.7 & 0.95\\
\midrule
Quadrotor & \miss, $N=8$ & 8 & 264 & 2.8 & 0.68\\
 & Select-then-filter & 8 & 33 & 3.5 & 0.69\\
 & MPPI (512 samples, horizon 100) & 0 & 51\,200 & 13.6 & 0.61\\
 & \miss, $N=256$ & 256 & 8\,448 & 57.1 & 0.58\\
 & \miss, $N=1024$ & 1024 & 33\,792 & 231.0 & 0.55\\
 & \miss-V, $L=16$ & $\le$2\,056 & $\le$71\,944 & 135.7 & 0.93\\
 & \miss-VL & 8 & 264 & 2.9 & \textbf{0.98}\\
\bottomrule\end{tabular}
}
\end{minipage}\par
\par\medskip\noindent\begin{minipage}{\textwidth}\heretable\centering
\caption{\miss-V ablation on the unicycle's unseen zones (task critic, three policies $\times$ 500 shared scenarios): lookahead $L$, continuations $M$, threshold $v_{\min}$, viability alone, \miss-VL and a success critic trained on the same logs (network cost on top of \miss). Viability is nearly always close to 0 or 1, so even $M=1$ separates the candidates; $L$ must cover the time to commit to a route.}
\label{tab:castvabl}
{\scriptsize\setlength{\tabcolsep}{5pt}\begin{tabular}{lcc}
\toprule
\miss-V variant (default $L=8$, $M=4$, $v_{\min}=0.5$) & Joint success & ms / decision \\
\midrule
$L=1$ & 0.823 [0.797, 0.848] & 6.0 \\
$L=2$ & 0.829 [0.804, 0.854] & 8.0 \\
$L=4$ & 0.803 [0.776, 0.830] & 13.3 \\
$L=8$ (default) & 0.993 [0.988, 0.997] & 18.4 \\
$M=1$ & 0.979 [0.970, 0.987] & 4.8 \\
$M=2$ & 0.971 [0.961, 0.981] & 7.7 \\
$M=8$ & \textbf{0.995} [0.990, 0.998] & 19.8 \\
$M=16$ & \textbf{0.995} [0.991, 0.999] & 33.9 \\
$v_{\min}=0.25$ & 0.971 [0.961, 0.980] & 11.9 \\
$v_{\min}=0.75$ & \textbf{0.995} [0.991, 0.999] & 12.2 \\
viability only ($L=8$) & 0.991 [0.983, 0.997] & 21.3 \\
\miss-VL: viability learned from logs & 0.949 [0.934, 0.962] & \miss${}+0.04$ \\
success critic from the same logs (no viability) & 0.967 [0.956, 0.978] & \miss${}+0.04$ \\
\midrule
\miss, task critic (reference) & 0.809 & 1.7 \\
\bottomrule
\end{tabular}
}
\end{minipage}\par

\par\medskip\noindent\begin{minipage}{\textwidth}\heretable\centering
\caption{The fallback, edit or certified backup, with the crossing pedestrian (three policies $\times$ 500 shared scenarios, the controllers of \figref{fig:ped}). ``Edit'': the output filter's correction, executed even if it does not end certified. ``Backup'': instead, the rest of the braking backup certified with the last certified chunk (\propref{prop:backupexo}). Left: the per-step forecast margins of \figref{fig:ped}; right: margins calibrated over whole episodes ($\varepsilon=0.05$), as \thmref{thm:safety}(b) assumes. No contact began in a certified chunk in any of the \BkEpisodesAll{} episodes. With edits, every contact began in an uncertified correction (\BkEdgeUncert{} of \BkEdgeN, per-step margins). With the backup and episode margins, no contact occurred during a certified step; all \BkAfterE{} came after the backup had brought the robot to rest. With per-step margins, which give no episode-level guarantee, \BkInS{} contacts fell inside the backup's certified steps. Joint success with the backup minus with edits, episode margins: filter \BkFilterGE, \miss \BkCastGE, \miss-V \BkCastvGE.}
\label{tab:backup}
{\scriptsize\setlength{\tabcolsep}{5pt}\begin{tabular}{llcccc}\toprule
 & & \multicolumn{2}{c}{per-step margins (Figure~\ref{fig:ped})} & \multicolumn{2}{c}{episode margins (\thmref{thm:safety}b)}\\
Controller & Fallback & Joint & Contact (\%) & Joint & Contact (\%)\\\midrule
Filter & edit & 0.72 & 1.5 & 0.61 & 0.9\\
 & backup & 0.84 & 3.5 & 0.76 & 2.0\\
\addlinespace[1pt]
\miss & edit & 0.73 & 2.4 & 0.60 & 0.7\\
 & backup & 0.87 & 3.7 & 0.81 & 1.3\\
\addlinespace[1pt]
\miss-V & edit & 0.82 & 1.0 & 0.91 & 0.0\\
 & backup & 0.91 & 3.3 & 0.93 & 0.0\\
\addlinespace[1pt]
\miss-VL & edit & 0.95 & 0.1 & 0.89 & 0.0\\
 & backup & 0.95 & 0.1 & 0.90 & 0.0\\
\addlinespace[1pt]
\miss, log critic & edit & 0.98 & 0.1 & 0.93 & 0.1\\
 & backup & 0.98 & 0.1 & 0.94 & 0.0\\
\bottomrule
\end{tabular}
}
\end{minipage}\par

\par\medskip\noindent\begin{minipage}{\textwidth}\heretable\centering
\caption{Planar quadrotor: joint success, mean of three policies (500 shared scenarios each); ``shuffled'': \miss with its critic's scores permuted. No certified controller entered a zone or exceeded a budget.}
\label{tab:quad}
{\scriptsize\setlength{\tabcolsep}{4pt}\begin{tabular}{lccccccc}
\toprule
Budget $b$ & filter & critic-free & sel.-then-filter & \miss & \miss + guard & shuffled & \miss-V \\
\midrule
1.01 & \textbf{0.17} & 0.02 & 0.11 & 0.00 & 0.00 & 0.03 & 0.01 \\
1.015 & 0.58 & 0.62 & 0.71 & 0.83 & 0.74 & 0.61 & \textbf{0.89} \\
1.02 & 0.59 & 0.77 & 0.83 & 0.97 & 0.96 & 0.74 & \textbf{0.99} \\
1.025 & 0.58 & 0.74 & 0.92 & 0.99 & 0.98 & 0.73 & \textbf{1.00} \\
1.03 & 0.58 & 0.72 & 0.97 & \textbf{1.00} & 0.99 & 0.71 & \textbf{1.00} \\
1.05 & 0.63 & 0.68 & \textbf{1.00} & \textbf{1.00} & \textbf{1.00} & 0.67 & \textbf{1.00} \\
\midrule
\multicolumn{8}{p{8.6cm}}{$b=1.02$, $N=1,2,4,8,16$: \miss 0.59, 0.91, 0.94, 0.97, 0.99; select-then-filter 0.59, 0.79, 0.84, 0.83, 0.86} \\
\midrule
\multicolumn{8}{p{8.6cm}}{Zone, joint success: frozen 0.36 (violates 40\,\%), filter 0.44, critic-free 0.43, MPPI 0.61} \\
\multicolumn{8}{p{8.6cm}}{task critic: sel.-then-filter 0.69, \miss 0.68, guard 0.70, shuffled 0.43; request-conditioned: \miss 1.00, sel.-then-filter 1.00} \\
\bottomrule
\end{tabular}
}
\end{minipage}\par

\par\medskip\noindent\begin{minipage}{\textwidth}\heretable\centering
\caption{Humanoid: joint success, mean of three policies (300 shared scenarios each). No certified controller exceeded the budget. $^\dagger$: budgets added after the planned run.}
\label{tab:humanoid}
{\scriptsize\setlength{\tabcolsep}{5pt}\begin{tabular}{lcccccc}
\toprule
Budget & filter & critic-free & sel.-then-filter & \miss & \miss + guard & \miss-V \\
\midrule
0.50$^\dagger$ & 0.00 & 0.00 & 0.04 & 0.01 & 0.00 & \textbf{0.16} \\
0.52$^\dagger$ & 0.00 & 0.00 & 0.16 & 0.09 & 0.01 & \textbf{0.51} \\
0.55 & 0.02 & 0.03 & 0.60 & 0.57 & 0.35 & \textbf{0.86} \\
0.58 & 0.18 & 0.25 & 0.82 & 0.82 & 0.79 & \textbf{0.87} \\
0.60 & 0.35 & 0.43 & 0.86 & 0.85 & \textbf{0.87} & 0.86 \\
0.65 & 0.48 & 0.51 & 0.88 & 0.88 & 0.88 & \textbf{0.89} \\
\midrule
\multicolumn{7}{l}{No request: frozen policy 0.49, critic best-of-8 0.88} \\
\bottomrule
\end{tabular}
}
\end{minipage}\par

\par\medskip\noindent\begin{minipage}{\textwidth}\heretable\centering
\caption{Unicycle with a keep-out zone: joint success, violation and stall rates, mean of three policies (500 scenarios each). Left: new zones of the family, unseen by the critics; right: one fixed zone. Matched critics differ only in the label. The planners do not use the policy (one set of 500 scenarios). Arms that draw extra samples use independent latents.}
\label{tab:zone}
{\scriptsize\setlength{\tabcolsep}{5pt}\begin{tabular}{lcccccc}
\toprule
 & \multicolumn{3}{c}{unseen zones from the family} & \multicolumn{3}{c}{fixed zone} \\
Controller & joint & violation & stall & joint & violation & stall \\
\midrule
Frozen policy & 0.479 & 0.503 & 0.000 & -- & -- & -- \\
Output CBF filter & 0.721 & 0.000 & 0.205 & 0.726 & 0.000 & 0.207 \\
Predictive safety filter & 0.746 & 0.000 & 0.157 & 0.732 & 0.000 & 0.182 \\
DPCC & 0.741 & 0.000 & 0.205 & 0.735 & 0.000 & 0.219 \\
Critic-free selection & 0.738 & 0.000 & 0.187 & -- & -- & -- \\
Select-then-filter, task critic & 0.860 & 0.000 & 0.113 & 0.867 & 0.000 & 0.115 \\
MPPI (knows model, map, goal) & 0.908 & 0.000 & 0.022 & 0.948 & 0.000 & 0.016 \\
MPC, hard constraints (same) & 0.900 & 0.000 & 0.072 & 0.922 & 0.000 & 0.058 \\
\midrule
\miss, task critic & 0.809 & 0.000 & 0.147 & 0.788 & 0.000 & 0.159 \\
\miss, matched task-target critic & 0.827 & 0.000 & 0.146 & 0.821 & 0.000 & 0.145 \\
\miss, request-conditioned critic & 0.984 & 0.000 & 0.001 & \textbf{0.987} & 0.000 & 0.006 \\
\midrule
\miss + guard, task critic & 0.786 & 0.000 & 0.161 & -- & -- & -- \\
\miss + guard, matched task-target critic & 0.828 & 0.000 & 0.143 & -- & -- & -- \\
\miss + guard, request-conditioned critic & 0.984 & 0.000 & 0.001 & -- & -- & -- \\
Select-then-filter, matched task-target critic & 0.877 & 0.000 & 0.087 & -- & -- & -- \\
Select-then-filter, request-conditioned critic & 0.984 & 0.000 & 0.001 & \textbf{0.987} & 0.000 & 0.003 \\
\midrule
\miss-V, task critic, $L=8$ & \textbf{0.993} & 0.000 & 0.004 & -- & -- & -- \\
\miss-V, task critic, $L=4$ & 0.803 & 0.000 & 0.173 & -- & -- & -- \\
\miss-V, task critic, $L=2$ & 0.829 & 0.000 & 0.151 & -- & -- & -- \\
\miss-V, task critic, $L=1$ & 0.823 & 0.000 & 0.153 & -- & -- & -- \\
Viability only, $L=8$ & 0.991 & 0.000 & 0.008 & -- & -- & -- \\
\bottomrule
\end{tabular}
}
\end{minipage}\par

\par\medskip\noindent\begin{minipage}{\textwidth}\heretable\centering
\caption{Can sweeps behind \figref{fig:regime}: joint success, 200 shared scenarios per cell; (v\,\%) marks violation rates of at least 0.5\,\%; at 9 and 12\,cm every controller violates, as empty batches become frequent (Safety paragraph of \secref{sec:quad}). ``Chunk cap'' caps every chunk; ``scaling'' is \miss with $N=1$. The guard column of the sample sweep, the scaling column and the last row were added after the planned run (best of eight without a request: \CdCanFreeGreedy{} on the scenarios of the ranking intervention).}
\label{tab:regime}
{\scriptsize\setlength{\tabcolsep}{3pt}\begin{tabular}{lccccccc}
\toprule
Obstacle radius (cm) & frozen & filter & DPCC & critic-free & sel.-then-filter & \miss & \miss + guard \\
\midrule
3 & 0.46 {\tiny(v18)} & 0.56 {\tiny(v1)} & 0.56 & 0.55 & 0.87 & 0.86 & \textbf{0.92} \\
6 & 0.34 {\tiny(v35)} & 0.56 {\tiny(v2)} & 0.58 {\tiny(v0.5)} & 0.57 {\tiny(v1)} & 0.85 & 0.87 & \textbf{0.90} {\tiny(v2)} \\
9 & 0.24 {\tiny(v52)} & 0.50 {\tiny(v6)} & 0.52 {\tiny(v8)} & 0.53 {\tiny(v6)} & 0.78 {\tiny(v6)} & 0.81 {\tiny(v6)} & \textbf{0.83} {\tiny(v6)} \\
12 & 0.11 {\tiny(v76)} & 0.34 {\tiny(v30)} & 0.36 {\tiny(v32)} & 0.34 {\tiny(v31)} & 0.51 {\tiny(v32)} & 0.55 {\tiny(v32)} & \textbf{0.58} {\tiny(v30)} \\
\midrule
Effort budget $b$ & & chunk cap & scaling & critic-free & sel.-then-scale & \miss & \miss + guard \\
\midrule
0.10 & & 0.09 & 0.10 & 0.08 & 0.07 & \textbf{0.28} & 0.27 \\
0.15 & & 0.23 & 0.26 & 0.19 & 0.16 & \textbf{0.53} & \textbf{0.53} \\
0.20 & & 0.29 & 0.32 & 0.23 & 0.31 & 0.64 & \textbf{0.67} \\
0.25 & & 0.37 & 0.46 & 0.35 & 0.43 & \textbf{0.71} & 0.68 \\
0.30 & & 0.41 & 0.45 & 0.44 & 0.63 & 0.73 & \textbf{0.79} \\
0.40 & & 0.42 & 0.51 & 0.47 & 0.77 & \textbf{0.84} & 0.81 \\
0.60 & & 0.53 & 0.52 & 0.54 & 0.87 & 0.88 & \textbf{0.91} \\
\midrule
Samples $N$ (obstacle 6 cm / budget 0.25) & & & & critic-free & sel.-then-filter & \miss & \miss + guard \\
\midrule
1 & & & & 0.56 / 0.46 & 0.56 / 0.46 & 0.56 / 0.46 & 0.56 / 0.46 \\
2 & & & & 0.56 / 0.39 & 0.85 / \textbf{0.70} & \textbf{0.87} / 0.66 & 0.83 / 0.60 \\
4 & & & & 0.56 / 0.38 & \textbf{0.90} / 0.69 & 0.89 / \textbf{0.75} & \textbf{0.90} / 0.71 \\
8 & & & & 0.57 / 0.35 & 0.85 / 0.43 & 0.87 / \textbf{0.71} & \textbf{0.90} / 0.68 \\
16 & & & & 0.56 / 0.33 & 0.80 / 0.41 & 0.80 / \textbf{0.68} & \textbf{0.86} / 0.67 \\
32 & & & & 0.57 / 0.29 & 0.74 / 0.35 & 0.77 / \textbf{0.66} & \textbf{0.78} / 0.60 \\
\midrule
\multicolumn{8}{l}{No request, critic's best of $N$ samples, $N=1,2,4,8,16,32$: 0.56, 0.87, 0.91, 0.86, 0.80, 0.77} \\
\bottomrule
\end{tabular}
}
\end{minipage}\par

\par\medskip\noindent\begin{minipage}{\textwidth}\heretable\centering
\caption{\miss-V without simulated continuations (unseen zones, 3 policies $\times$ 500 shared scenarios). Model: simulated continuations. Logs: an ensemble trained on 3000 logged episodes per policy of the critic-free certified controller, labels in hindsight (held-out AUC $\ge\MfAucLo$). Gain $g$: execution scales the commands, predictions keep the nominal model. Logs minus model under mismatch: \MfMmGap{} (pooled over both gains); violations at $g=0.7$ (worst policy; \miss, \miss-V, \miss-VL): \MfViolCast{}\,\%, \MfViolMb{}\,\%, \MfViolMf{}\,\%. \propref{prop:est} does not cover the learned estimate.}
\label{tab:modelfree}
{\scriptsize\setlength{\tabcolsep}{5pt}\begin{tabular}{lccc}\toprule
Unseen zones & \miss & \miss-V, model & \miss-VL\\\midrule
Unicycle, exact model & 0.81 & \textbf{0.99} & 0.95\\
Unicycle, actuator gain 0.85 & 0.70 & \textbf{0.96} & 0.93\\
Unicycle, actuator gain 0.7 & 0.65 & 0.89 & \textbf{0.90}\\
Quadrotor, exact model & 0.68 & 0.93 & \textbf{0.98}\\
\bottomrule\end{tabular}
}
\end{minipage}\par

\textbf{Calibrated forecast margins on Can.} \tabref{tab:calib} checks \thmref{thm:safety}(b) on Can with a slow obstacle seen through a tracker with 5\,mm noise. Miscoverage is the share of episodes in which a forecast left its margin; $\hat\Prob[\mathcal B]$ is the share in which a hand-prediction error exceeded the certificate's slack. With margins calibrated over whole episodes, the closed-loop miscoverage stays at or below its target $\varepsilon$ up to sampling error (largest excess: 0.055 against 0.05) for the filter and for \miss alike, as the theorem states, and collisions stay at or below 0.5\,\%. Per-step margins give no episode-level guarantee: a forecast leaves them in more than half of the episodes.

\par\medskip\noindent\begin{minipage}{\textwidth}\heretable\centering
\caption{Episode-level calibrated safety on Can (200 episodes per cell; \miss with the unanimity guard). $s_\varepsilon$: margin factor over the per-step 0.99 margins. $^\dagger$The per-step margins that apply after the first decision, the reference for $s_\varepsilon$.}
\label{tab:calib}
{\scriptsize\setlength{\tabcolsep}{3pt}\begin{tabular}{lccccccc}
\toprule
Margins & $s_\varepsilon$ & miscoverage & miscoverage & collision & $\hat\Prob[\mathcal B]$ & joint success & \miss{} $-$ filter \\
 & & offline, all & closed loop & & & & \\
 & & decisions & \miss{} / filter & \miss{} / filter & \miss{} / filter & \miss{} / filter & \\
\midrule
$\varepsilon=0.5$ & 1.07 & 0.505 & 0.400 / 0.415 & 0.000 / 0.005 & 0.080 / 0.045 & 0.69 / 0.43 & +0.27 {\tiny[+0.18,+0.34]} \\
$\varepsilon=0.2$ & 1.17 & 0.185 & 0.165 / 0.185 & 0.000 / 0.000 & 0.090 / 0.090 & 0.67 / 0.39 & +0.28 {\tiny[+0.20,+0.36]} \\
$\varepsilon=0.05$ & 1.30 & 0.041 & 0.045 / 0.055 & 0.000 / 0.000 & 0.120 / 0.110 & 0.64 / 0.38 & +0.26 {\tiny[+0.18,+0.34]} \\
per step 0.99$^\dagger$ & 1.00 & 0.733 & 0.565 / 0.625 & 0.000 / 0.000 & 0.055 / 0.030 & 0.70 / 0.47 & +0.23 {\tiny[+0.15,+0.31]} \\
\bottomrule
\end{tabular}
}
\end{minipage}\par

\end{document}